\documentclass[11pt]{article}

\usepackage[letterpaper,margin=0.9in,headheight=14pt]{geometry}
\usepackage[utf8]{inputenc}
\usepackage{xcolor}
\definecolor{MidnightBlue}{HTML}{12233F}
\definecolor{RadarBlue}{HTML}{2563EB}
\definecolor{RadarInk}{HTML}{243247}
\definecolor{RadarMuted}{HTML}{5C6B80}
\definecolor{RadarRule}{HTML}{D7DFEA}
\definecolor{RadarSky}{HTML}{EAF1FF}
\definecolor{LinkRed}{HTML}{CC0000}
\definecolor{CiteBlue}{HTML}{0000CC}
\usepackage{graphicx}
\usepackage{pgf}
\usepackage{tikz}
\usepackage{helvet}
\usepackage{array}
\usepackage{booktabs}
\usepackage{tabularx}
\usepackage{longtable}
\usepackage[numbers,sort&compress]{natbib}
\usepackage[hypcap=false]{caption}
\usepackage[section]{placeins}
\usepackage{authblk}
\usepackage[bottom]{footmisc}
\usepackage{hyperref}
\usepackage{fancyhdr}

\hypersetup{
  colorlinks=true,
  linkcolor=LinkRed,
  urlcolor=CiteBlue,
  citecolor=CiteBlue,
  pdftitle={Benchmark Radar: A Living Database and Search Engine for AI Benchmarks and Evaluation},
  pdfauthor={Koutian Wu, Wenjun Zhang, Junjie Zhou, Pengqian Han, Qiming Bao, Ergan Shang, Jiayu Wang, Junkai Wang, Wanghan Xu, Songyuanyi Lu, Yixuan Li, Rudy Mizrahi Celekli, Lazizbek Ravshanov, Yaoliang Bian, Xudong Han, and Lin Shi}
}

\renewcommand{\arraystretch}{1.15}
\newcolumntype{Y}{>{\raggedright\arraybackslash}X}
\newcolumntype{L}[1]{>{\raggedright\arraybackslash}p{#1}}
\usepackage{pifont}
\newcommand{\cmark}{\ding{51}}
\newcommand{\xmark}{\ding{55}}
\newcommand{\ReportDataCutoff}{2026-09-07}
\newcommand{\ReportObservationCount}{11068}
\newcommand{\ReportArtifactCount}{6546}
\newcommand{\ReportSnapshotCount}{46}
\newcommand{\ReportCatalogCount}{1283}
\newcommand{\ReportCatalogSourceCount}{4}
\newcommand{\ReportConnectorCount}{13}
\newcommand{\ReportFirstPartyFeedCount}{24}
\newcommand{\ReportIngestSourceCount}{37}
\newcommand{\ReportLLMStatsCount}{687}
\newcommand{\ReportOpenCompassCount}{461}
\newcommand{\ReportArtificialAnalysisCount}{25}
\newcommand{\ReportModelReportCount}{110}

\newcommand{\CensusRecords}{1283}
\newcommand{\CensusNumericScores}{12916}
\newcommand{\CensusScored}{790}
\newcommand{\CensusUnscored}{493}
\newcommand{\CensusOtherNumeric}{708}
\newcommand{\CensusDocumentsKnown}{1278}

\newcommand{\CensusReleaseKnown}{615}
\newcommand{\CensusPaperLinks}{475}
\newcommand{\CensusRepoLinks}{506}
\newcommand{\CensusDatasetLinks}{293}
\newcommand{\CensusNoScoreWithLinks}{464}
\newcommand{\CensusPercent}{82}
\newcommand{\CensusDocuments}{1208}
\newcommand{\CensusModels}{868}

\newcommand{\CensusSimulatedSnapshots}{4}

\newcommand{\CensusMultisourceArtifacts}{59}
\newcommand{\CensusUnclassified}{5863}
\newcommand{\CensusSourceRows}{%
LLM Stats & 687 & 679 & 8 & 5,544\\
OpenCompass Hub & 461 & 0 & 461 & 0\\
Artificial Analysis & 25 & 25 & 0 & 7,050\\
Model reports & 110 & 86 & 24 & 322\\
}
\newcommand{\CensusMarks}{%
\CensusDot{0}{0}{0}{RadarBlue}{llm-stats-aa-briefcase}%
\CensusDot{0}{1}{0}{RadarBlue}{llm-stats-aa-index}%
\CensusDot{0}{2}{0}{RadarBlue}{llm-stats-aa-lcr}%
\CensusDot{0}{3}{0}{RadarBlue}{llm-stats-aa-omniscience-index}%
\CensusDot{0}{4}{0}{RadarBlue}{llm-stats-acebench}%
\CensusDot{0}{5}{0}{RadarBlue}{llm-stats-activitynet}%
\CensusDot{0}{6}{0}{RadarBlue}{llm-stats-advancedif}%
\CensusDot{0}{7}{0}{RadarBlue}{llm-stats-aethercode}%
\CensusDot{0}{8}{0}{RadarBlue}{llm-stats-agent-startup-bench}%
\CensusDot{0}{9}{0}{RadarBlue}{llm-stats-agents-last-exam}%
\CensusDot{0}{10}{0}{RadarBlue}{llm-stats-agieval}%
\CensusDot{0}{11}{0}{RadarBlue}{llm-stats-ai2-reasoning-challenge-arc}%
\CensusDot{0}{12}{0}{RadarBlue}{llm-stats-ai2d}%
\CensusDot{0}{13}{0}{RadarBlue}{llm-stats-aider}%
\CensusDot{0}{14}{0}{RadarBlue}{llm-stats-aider-polyglot}%
\CensusDot{0}{15}{0}{RadarBlue}{llm-stats-aider-polyglot-edit}%
\CensusDot{0}{16}{0}{RadarBlue}{llm-stats-aime}%
\CensusDot{0}{17}{0}{RadarBlue}{llm-stats-aime-2024}%
\CensusDot{0}{18}{0}{RadarBlue}{llm-stats-aime-2025}%
\CensusDot{0}{19}{0}{RadarBlue}{llm-stats-aime-2026}%
\CensusDot{0}{0}{1}{RadarBlue}{llm-stats-air-bench}%
\CensusDot{0}{1}{1}{RadarBlue}{llm-stats-aitz-em}%
\CensusDot{0}{2}{1}{RadarBlue}{llm-stats-alignbench}%
\CensusDot{0}{3}{1}{RadarBlue}{llm-stats-alpacaeval-2-0}%
\CensusDot{0}{4}{1}{RadarBlue}{llm-stats-amc-2022-23}%
\CensusDot{0}{5}{1}{RadarBlue}{llm-stats-amo-bench}%
\CensusDot{0}{6}{1}{RadarBlue}{llm-stats-android-control-high-em}%
\CensusDot{0}{7}{1}{RadarBlue}{llm-stats-android-control-low-em}%
\CensusDot{0}{8}{1}{RadarBlue}{llm-stats-androidbench}%
\CensusDot{0}{9}{1}{RadarBlue}{llm-stats-androidworld}%
\CensusDot{0}{10}{1}{RadarBlue}{llm-stats-androidworld-sr}%
\CensusDot{0}{11}{1}{RadarBlue}{llm-stats-apex}%
\CensusDot{0}{12}{1}{RadarBlue}{llm-stats-apex-agents}%
\CensusDot{0}{13}{1}{RadarBlue}{llm-stats-apex-swe}%
\CensusDot{0}{14}{1}{RadarBlue}{llm-stats-api-bank}%
\CensusDot{0}{15}{1}{RadarBlue}{llm-stats-arc}%
\CensusDot{0}{16}{1}{RadarBlue}{llm-stats-arc-agi}%
\CensusDot{0}{17}{1}{RadarBlue}{llm-stats-arc-agi-3}%
\CensusDot{0}{18}{1}{RadarBlue}{llm-stats-arc-agi-v2}%
\CensusDot{0}{19}{1}{RadarBlue}{llm-stats-arc-c}%
\CensusDot{0}{0}{2}{RadarBlue}{llm-stats-arc-e}%
\CensusDot{0}{1}{2}{RadarBlue}{llm-stats-arcagi2}%
\CensusDot{0}{2}{2}{RadarBlue}{llm-stats-arena-hard}%
\CensusDot{0}{3}{2}{RadarBlue}{llm-stats-arena-hard-v2}%
\CensusDot{0}{4}{2}{RadarBlue}{llm-stats-arkitscenes}%
\CensusDot{0}{5}{2}{RadarBlue}{llm-stats-artifacts-bench}%
\CensusDot{0}{6}{2}{RadarBlue}{llm-stats-artificial-analysis}%
\CensusDot{0}{7}{2}{RadarBlue}{llm-stats-arxivmath}%
\CensusDot{0}{8}{2}{RadarBlue}{llm-stats-attaq}%
\CensusDot{0}{9}{2}{RadarBlue}{llm-stats-autologi}%
\CensusDot{0}{10}{2}{RadarBlue}{llm-stats-automationbench}%
\CensusDot{0}{11}{2}{RadarBlue}{llm-stats-automationbench-aa}%
\CensusDot{0}{12}{2}{RadarBlue}{llm-stats-babyvision}%
\CensusDot{0}{13}{2}{RadarBlue}{llm-stats-bankertoolbench}%
\CensusDot{0}{14}{2}{RadarBlue}{llm-stats-bbh}%
\CensusDot{0}{15}{2}{RadarBlue}{llm-stats-bc-vl}%
\CensusDot{0}{16}{2}{RadarBlue}{llm-stats-beam-128k}%
\CensusDot{0}{17}{2}{RadarBlue}{llm-stats-benchcad}%
\CensusDot{0}{18}{2}{RadarBlue}{llm-stats-benchcad-with-python-tool}%
\CensusDot{0}{19}{2}{RadarBlue}{llm-stats-beyond-aime}%
\CensusDot{0}{0}{3}{RadarBlue}{llm-stats-bfcl}%
\CensusDot{0}{1}{3}{RadarBlue}{llm-stats-bfcl-v2}%
\CensusDot{0}{2}{3}{RadarBlue}{llm-stats-bfcl-v3}%
\CensusDot{0}{3}{3}{RadarBlue}{llm-stats-bfcl-v3-multiturn}%
\CensusDot{0}{4}{3}{RadarBlue}{llm-stats-bfcl-v4}%
\CensusDot{0}{5}{3}{RadarBlue}{llm-stats-big-bench}%
\CensusDot{0}{6}{3}{RadarBlue}{llm-stats-big-bench-audio}%
\CensusDot{0}{7}{3}{RadarBlue}{llm-stats-big-bench-extra-hard}%
\CensusDot{0}{8}{3}{RadarBlue}{llm-stats-big-bench-hard}%
\CensusDot{0}{9}{3}{RadarBlue}{llm-stats-big-finance-bench}%
\CensusDot{0}{10}{3}{RadarBlue}{llm-stats-bigcodebench}%
\CensusDot{0}{11}{3}{RadarBlue}{llm-stats-bigcodebench-full}%
\CensusDot{0}{12}{3}{RadarBlue}{llm-stats-bigcodebench-hard}%
\CensusDot{0}{13}{3}{RadarBlue}{llm-stats-biolp-bench}%
\CensusDot{0}{14}{3}{RadarBlue}{llm-stats-biomysterybench}%
\CensusDot{0}{15}{3}{RadarBlue}{llm-stats-bird-sql-dev}%
\CensusDot{0}{16}{3}{RadarBlue}{llm-stats-bixbench}%
\CensusDot{0}{17}{3}{RadarBlue}{llm-stats-blink}%
\CensusDot{0}{18}{3}{RadarBlue}{llm-stats-blueprint-bench-2}%
\CensusDot{0}{19}{3}{RadarBlue}{llm-stats-boolq}%
\CensusDot{0}{0}{4}{RadarBlue}{llm-stats-browsecomp}%
\CensusDot{0}{1}{4}{RadarBlue}{llm-stats-browsecomp-long-128k}%
\CensusDot{0}{2}{4}{RadarBlue}{llm-stats-browsecomp-long-256k}%
\CensusDot{0}{3}{4}{RadarBlue}{llm-stats-browsecomp-vl}%
\CensusDot{0}{4}{4}{RadarBlue}{llm-stats-browsecomp-zh}%
\CensusDot{0}{5}{4}{RadarBlue}{llm-stats-c-eval}%
\CensusDot{0}{6}{4}{RadarBlue}{llm-stats-capture-the-flag-challenges}%
\CensusDot{0}{7}{4}{RadarBlue}{llm-stats-cbnsl}%
\CensusDot{0}{8}{4}{RadarBlue}{llm-stats-cc-bench-v2-backend}%
\CensusDot{0}{9}{4}{RadarBlue}{llm-stats-cc-bench-v2-frontend}%
\CensusDot{0}{10}{4}{RadarBlue}{llm-stats-cc-bench-v2-repo}%
\CensusDot{0}{11}{4}{RadarBlue}{llm-stats-cc-ocr}%
\CensusDot{0}{12}{4}{RadarBlue}{llm-stats-cfeval}%
\CensusDot{0}{13}{4}{RadarBlue}{llm-stats-charadessta}%
\CensusDot{0}{14}{4}{RadarBlue}{llm-stats-chartmuseum}%
\CensusDot{0}{15}{4}{RadarBlue}{llm-stats-chartqa}%
\CensusDot{0}{16}{4}{RadarBlue}{llm-stats-chartqapro}%
\CensusDot{0}{17}{4}{RadarBlue}{llm-stats-charxiv-d}%
\CensusDot{0}{18}{4}{RadarBlue}{llm-stats-charxiv-r}%
\CensusDot{0}{19}{4}{RadarBlue}{llm-stats-chexpert-cxr}%
\CensusDot{0}{0}{5}{RadarBlue}{llm-stats-ci-memories-coverage}%
\CensusDot{0}{1}{5}{RadarBlue}{llm-stats-ci-memories-violation}%
\CensusDot{0}{2}{5}{RadarBlue}{llm-stats-cl-bench}%
\CensusDot{0}{3}{5}{RadarBlue}{llm-stats-cl-bench-life}%
\CensusDot{0}{4}{5}{RadarBlue}{llm-stats-claw-eval}%
\CensusDot{0}{5}{5}{RadarBlue}{llm-stats-claw-eval-mm}%
\CensusDot{0}{6}{5}{RadarBlue}{llm-stats-cloningscenarios}%
\CensusDot{0}{7}{5}{RadarBlue}{llm-stats-cluewsc}%
\CensusDot{0}{8}{5}{RadarBlue}{llm-stats-cmmlu}%
\CensusDot{0}{9}{5}{RadarBlue}{llm-stats-cmt-benchmark}%
\CensusDot{0}{10}{5}{RadarBlue}{llm-stats-cnmo-2024}%
\CensusDot{0}{11}{5}{RadarBlue}{llm-stats-codeforces}%
\CensusDot{0}{12}{5}{RadarBlue}{llm-stats-codegolf-v2-2}%
\CensusDot{0}{13}{5}{RadarBlue}{llm-stats-cohere-agentic-question-answering}%
\CensusDot{0}{14}{5}{RadarBlue}{llm-stats-cohere-data-analysis}%
\CensusDot{0}{15}{5}{RadarBlue}{llm-stats-cohere-memory-usage-quality}%
\CensusDot{0}{16}{5}{RadarBlue}{llm-stats-collie}%
\CensusDot{0}{17}{5}{RadarBlue}{llm-stats-common-voice-15}%
\CensusDot{0}{18}{5}{RadarBlue}{llm-stats-commonsenseqa}%
\CensusDot{0}{19}{5}{RadarGray}{llm-stats-community-07c9946d-dcf0-4977-a640-a6b1356b4f0b}%
\CensusDot{0}{0}{6}{RadarGray}{llm-stats-community-2256e9c9-b256-4444-b639-7cc3b1855d96}%
\CensusDot{0}{1}{6}{RadarGray}{llm-stats-community-5f95f778-c521-43fa-b80e-6a55465601e3}%
\CensusDot{0}{2}{6}{RadarGray}{llm-stats-community-64d67847-06bd-423a-923c-c2acfab82281}%
\CensusDot{0}{3}{6}{RadarGray}{llm-stats-community-ed90e889-4678-4fbd-98ab-0e654f4bf35e}%
\CensusDot{0}{4}{6}{RadarGray}{llm-stats-community-fd462fc2-283c-4967-bd7d-b39d7c661807}%
\CensusDot{0}{5}{6}{RadarBlue}{llm-stats-complexfuncbench}%
\CensusDot{0}{6}{6}{RadarBlue}{llm-stats-contphy}%
\CensusDot{0}{7}{6}{RadarBlue}{llm-stats-corpusqa}%
\CensusDot{0}{8}{6}{RadarBlue}{llm-stats-corpusqa-1m}%
\CensusDot{0}{9}{6}{RadarBlue}{llm-stats-countbench}%
\CensusDot{0}{10}{6}{RadarBlue}{llm-stats-countqa}%
\CensusDot{0}{11}{6}{RadarBlue}{llm-stats-covost2}%
\CensusDot{0}{12}{6}{RadarBlue}{llm-stats-covost2-en-zh}%
\CensusDot{0}{13}{6}{RadarBlue}{llm-stats-coworkbench}%
\CensusDot{0}{14}{6}{RadarBlue}{llm-stats-crag}%
\CensusDot{0}{15}{6}{RadarBlue}{llm-stats-creative-writing-v3}%
\CensusDot{0}{16}{6}{RadarBlue}{llm-stats-creativework}%
\CensusDot{0}{17}{6}{RadarBlue}{llm-stats-critpt}%
\CensusDot{0}{18}{6}{RadarBlue}{llm-stats-crossvid}%
\CensusDot{0}{19}{6}{RadarBlue}{llm-stats-crperelation}%
\CensusDot{0}{0}{7}{RadarBlue}{llm-stats-crux-o}%
\CensusDot{0}{1}{7}{RadarBlue}{llm-stats-cruxeval-input-cot}%
\CensusDot{0}{2}{7}{RadarBlue}{llm-stats-cruxeval-o}%
\CensusDot{0}{3}{7}{RadarBlue}{llm-stats-cruxeval-output-cot}%
\CensusDot{0}{4}{7}{RadarBlue}{llm-stats-csimpleqa}%
\CensusDot{0}{5}{7}{RadarBlue}{llm-stats-cursorbench-3-2}%
\CensusDot{0}{6}{7}{RadarGray}{llm-stats-cvtg-2k}%
\CensusDot{0}{7}{7}{RadarBlue}{llm-stats-cybench}%
\CensusDot{0}{8}{7}{RadarBlue}{llm-stats-cybergym}%
\CensusDot{0}{9}{7}{RadarBlue}{llm-stats-cyberseceval-4}%
\CensusDot{0}{10}{7}{RadarBlue}{llm-stats-cybersecurity-ctfs}%
\CensusDot{0}{11}{7}{RadarBlue}{llm-stats-dailyomni}%
\CensusDot{0}{12}{7}{RadarBlue}{llm-stats-deck-bench}%
\CensusDot{0}{13}{7}{RadarBlue}{llm-stats-deep-planning}%
\CensusDot{0}{14}{7}{RadarBlue}{llm-stats-deepsearchqa}%
\CensusDot{0}{15}{7}{RadarBlue}{llm-stats-deepswe}%
\CensusDot{0}{16}{7}{RadarBlue}{llm-stats-deepswe-1-0}%
\CensusDot{0}{17}{7}{RadarBlue}{llm-stats-deepswe-1-1}%
\CensusDot{0}{18}{7}{RadarBlue}{llm-stats-dermmcqa}%
\CensusDot{0}{19}{7}{RadarBlue}{llm-stats-design2code}%
\CensusDot{0}{0}{8}{RadarBlue}{llm-stats-docvqa}%
\CensusDot{0}{1}{8}{RadarBlue}{llm-stats-docvqatest}%
\CensusDot{0}{2}{8}{RadarBlue}{llm-stats-doubao-multi-turn-bench}%
\CensusDot{0}{3}{8}{RadarBlue}{llm-stats-draco}%
\CensusDot{0}{4}{8}{RadarBlue}{llm-stats-drop}%
\CensusDot{0}{5}{8}{RadarBlue}{llm-stats-ds-arena-code}%
\CensusDot{0}{6}{8}{RadarBlue}{llm-stats-ds-fim-eval}%
\CensusDot{0}{7}{8}{RadarBlue}{llm-stats-dsbench-fullstack}%
\CensusDot{0}{8}{8}{RadarBlue}{llm-stats-dsbench-hard}%
\CensusDot{0}{9}{8}{RadarBlue}{llm-stats-dude}%
\CensusDot{0}{10}{8}{RadarBlue}{llm-stats-dynamath}%
\CensusDot{0}{11}{8}{RadarBlue}{llm-stats-eclektic}%
\CensusDot{0}{12}{8}{RadarBlue}{llm-stats-egoschema}%
\CensusDot{0}{13}{8}{RadarBlue}{llm-stats-embspatialbench}%
\CensusDot{0}{14}{8}{RadarBlue}{llm-stats-emma}%
\CensusDot{0}{15}{8}{RadarBlue}{llm-stats-eq-bench}%
\CensusDot{0}{16}{8}{RadarBlue}{llm-stats-erqa}%
\CensusDot{0}{17}{8}{RadarBlue}{llm-stats-evalplus}%
\CensusDot{0}{18}{8}{RadarBlue}{llm-stats-exploitbench}%
\CensusDot{0}{19}{8}{RadarBlue}{llm-stats-exploitgym}%
\CensusDot{0}{0}{9}{RadarBlue}{llm-stats-facts-grounding}%
\CensusDot{0}{1}{9}{RadarBlue}{llm-stats-factscore}%
\CensusDot{0}{2}{9}{RadarBlue}{llm-stats-figqa}%
\CensusDot{0}{3}{9}{RadarBlue}{llm-stats-finance-agent}%
\CensusDot{0}{4}{9}{RadarBlue}{llm-stats-finance-agent-v1-1}%
\CensusDot{0}{5}{9}{RadarBlue}{llm-stats-finance-agent-v2}%
\CensusDot{0}{6}{9}{RadarBlue}{llm-stats-finqa}%
\CensusDot{0}{7}{9}{RadarBlue}{llm-stats-finsearchcomp-t2-t3}%
\CensusDot{0}{8}{9}{RadarBlue}{llm-stats-finsearchcomp-t3}%
\CensusDot{0}{9}{9}{RadarBlue}{llm-stats-flame-vlm-code}%
\CensusDot{0}{10}{9}{RadarBlue}{llm-stats-flenqa}%
\CensusDot{0}{11}{9}{RadarBlue}{llm-stats-fleurs}%
\CensusDot{0}{12}{9}{RadarBlue}{llm-stats-frames}%
\CensusDot{0}{13}{9}{RadarBlue}{llm-stats-french-mmlu}%
\CensusDot{0}{14}{9}{RadarBlue}{llm-stats-frontier-bench-v0-1}%
\CensusDot{0}{15}{9}{RadarBlue}{llm-stats-frontier-science}%
\CensusDot{0}{16}{9}{RadarBlue}{llm-stats-frontier-swe-impl}%
\CensusDot{0}{17}{9}{RadarBlue}{llm-stats-frontiercode}%
\CensusDot{0}{18}{9}{RadarBlue}{llm-stats-frontiercode-1-1}%
\CensusDot{0}{19}{9}{RadarBlue}{llm-stats-frontiercs}%
\CensusDot{0}{0}{10}{RadarBlue}{llm-stats-frontiermath}%
\CensusDot{0}{1}{10}{RadarBlue}{llm-stats-frontiermath-tier-4-v2}%
\CensusDot{0}{2}{10}{RadarBlue}{llm-stats-frontierscience-olympiad}%
\CensusDot{0}{3}{10}{RadarBlue}{llm-stats-frontierscience-research}%
\CensusDot{0}{4}{10}{RadarBlue}{llm-stats-frontierswe}%
\CensusDot{0}{5}{10}{RadarBlue}{llm-stats-fullstackbench-en}%
\CensusDot{0}{6}{10}{RadarBlue}{llm-stats-fullstackbench-zh}%
\CensusDot{0}{7}{10}{RadarBlue}{llm-stats-functionalmath}%
\CensusDot{0}{8}{10}{RadarBlue}{llm-stats-gaia2}%
\CensusDot{0}{9}{10}{RadarBlue}{llm-stats-gameworld}%
\CensusDot{0}{10}{10}{RadarBlue}{llm-stats-gdp-pdf}%
\CensusDot{0}{11}{10}{RadarBlue}{llm-stats-gdpval}%
\CensusDot{0}{12}{10}{RadarBlue}{llm-stats-gdpval-aa}%
\CensusDot{0}{13}{10}{RadarBlue}{llm-stats-gdpval-mm}%
\CensusDot{0}{14}{10}{RadarBlue}{llm-stats-gdpval-rubrics}%
\CensusDot{0}{15}{10}{RadarBlue}{llm-stats-genebench}%
\CensusDot{0}{16}{10}{RadarBlue}{llm-stats-genebench-pro}%
\CensusDot{0}{17}{10}{RadarBlue}{llm-stats-giantsteps-tempo}%
\CensusDot{0}{18}{10}{RadarBlue}{llm-stats-global-mmlu}%
\CensusDot{0}{19}{10}{RadarBlue}{llm-stats-global-mmlu-lite}%
\CensusDot{0}{0}{11}{RadarBlue}{llm-stats-global-piqa}%
\CensusDot{0}{1}{11}{RadarBlue}{llm-stats-gorilla-benchmark-api-bench}%
\CensusDot{0}{2}{11}{RadarBlue}{llm-stats-govreport}%
\CensusDot{0}{3}{11}{RadarBlue}{llm-stats-gpqa}%
\CensusDot{0}{4}{11}{RadarBlue}{llm-stats-gpqa-biology}%
\CensusDot{0}{5}{11}{RadarBlue}{llm-stats-gpqa-chemistry}%
\CensusDot{0}{6}{11}{RadarBlue}{llm-stats-gpqa-physics}%
\CensusDot{0}{7}{11}{RadarBlue}{llm-stats-graphwalks}%
\CensusDot{0}{8}{11}{RadarBlue}{llm-stats-graphwalks-bfs-1m}%
\CensusDot{0}{9}{11}{RadarBlue}{llm-stats-graphwalks-bfs-128k}%
\CensusDot{0}{10}{11}{RadarBlue}{llm-stats-graphwalks-bfs-128k-2}%
\CensusDot{0}{11}{11}{RadarBlue}{llm-stats-graphwalks-parents-128k}%
\CensusDot{0}{12}{11}{RadarBlue}{llm-stats-graphwalks-parents-128k-2}%
\CensusDot{0}{13}{11}{RadarBlue}{llm-stats-groundui-1k}%
\CensusDot{0}{14}{11}{RadarBlue}{llm-stats-gsm-8k-cot}%
\CensusDot{0}{15}{11}{RadarBlue}{llm-stats-gsm8k}%
\CensusDot{0}{16}{11}{RadarBlue}{llm-stats-gsm8k-chat}%
\CensusDot{0}{17}{11}{RadarBlue}{llm-stats-hallusion-bench}%
\CensusDot{0}{18}{11}{RadarBlue}{llm-stats-harvey-lab}%
\CensusDot{0}{19}{11}{RadarBlue}{llm-stats-harvey-lab-aa}%
\CensusDot{0}{0}{12}{RadarBlue}{llm-stats-healthbench}%
\CensusDot{0}{1}{12}{RadarBlue}{llm-stats-healthbench-consensus}%
\CensusDot{0}{2}{12}{RadarBlue}{llm-stats-healthbench-hard}%
\CensusDot{0}{3}{12}{RadarBlue}{llm-stats-healthbench-professional}%
\CensusDot{0}{4}{12}{RadarBlue}{llm-stats-hellaswag}%
\CensusDot{0}{5}{12}{RadarBlue}{llm-stats-hiddenmath}%
\CensusDot{0}{6}{12}{RadarBlue}{llm-stats-hipho}%
\CensusDot{0}{7}{12}{RadarBlue}{llm-stats-hle-verified}%
\CensusDot{0}{8}{12}{RadarBlue}{llm-stats-hmmt-2025}%
\CensusDot{0}{9}{12}{RadarBlue}{llm-stats-hmmt-feb-26}%
\CensusDot{0}{10}{12}{RadarBlue}{llm-stats-hmmt25}%
\CensusDot{0}{11}{12}{RadarBlue}{llm-stats-horizonmath}%
\CensusDot{0}{12}{12}{RadarBlue}{llm-stats-hr-bench-4k}%
\CensusDot{0}{13}{12}{RadarBlue}{llm-stats-humaneval}%
\CensusDot{0}{14}{12}{RadarBlue}{llm-stats-humaneval-2}%
\CensusDot{0}{15}{12}{RadarBlue}{llm-stats-humaneval-average}%
\CensusDot{0}{16}{12}{RadarBlue}{llm-stats-humaneval-er}%
\CensusDot{0}{17}{12}{RadarBlue}{llm-stats-humaneval-mul}%
\CensusDot{0}{18}{12}{RadarBlue}{llm-stats-humaneval-plus}%
\CensusDot{0}{19}{12}{RadarBlue}{llm-stats-humanevalfim-average}%
\CensusDot{0}{0}{13}{RadarBlue}{llm-stats-humanity-s-last-exam}%
\CensusDot{0}{1}{13}{RadarBlue}{llm-stats-humanity-s-last-exam-no-tools-text-only}%
\CensusDot{0}{2}{13}{RadarBlue}{llm-stats-humanity-s-last-exam-with-tools-text-only}%
\CensusDot{0}{3}{13}{RadarBlue}{llm-stats-hypersim}%
\CensusDot{0}{4}{13}{RadarBlue}{llm-stats-if}%
\CensusDot{0}{5}{13}{RadarBlue}{llm-stats-ifbench}%
\CensusDot{0}{6}{13}{RadarBlue}{llm-stats-ifeval}%
\CensusDot{0}{7}{13}{RadarBlue}{llm-stats-image2floorplan}%
\CensusDot{0}{8}{13}{RadarBlue}{llm-stats-imagemining}%
\CensusDot{0}{9}{13}{RadarBlue}{llm-stats-imo-2025}%
\CensusDot{0}{10}{13}{RadarBlue}{llm-stats-imo-answerbench}%
\CensusDot{0}{11}{13}{RadarBlue}{llm-stats-imoproof-adv}%
\CensusDot{0}{12}{13}{RadarBlue}{llm-stats-include}%
\CensusDot{0}{13}{13}{RadarBlue}{llm-stats-infinitebench-en-mc}%
\CensusDot{0}{14}{13}{RadarBlue}{llm-stats-infinitebench-en-qa}%
\CensusDot{0}{15}{13}{RadarBlue}{llm-stats-infographicsqa}%
\CensusDot{0}{16}{13}{RadarBlue}{llm-stats-infovqa}%
\CensusDot{0}{17}{13}{RadarBlue}{llm-stats-infovqatest}%
\CensusDot{0}{18}{13}{RadarBlue}{llm-stats-instruct-humaneval}%
\CensusDot{0}{19}{13}{RadarBlue}{llm-stats-intergps}%
\CensusDot{0}{0}{14}{RadarBlue}{llm-stats-internal-api-instruction-following-hard}%
\CensusDot{0}{1}{14}{RadarBlue}{llm-stats-internal-research-debugging-evaluation}%
\CensusDot{0}{2}{14}{RadarBlue}{llm-stats-ipho-2025}%
\CensusDot{0}{3}{14}{RadarBlue}{llm-stats-job-bench}%
\CensusDot{0}{4}{14}{RadarBlue}{llm-stats-kernel-bench-l3}%
\CensusDot{0}{5}{14}{RadarBlue}{llm-stats-kernelbench-hard}%
\CensusDot{0}{6}{14}{RadarBlue}{llm-stats-kernelgen-1p}%
\CensusDot{0}{7}{14}{RadarBlue}{llm-stats-kimi-claw-24-7-bench}%
\CensusDot{0}{8}{14}{RadarBlue}{llm-stats-kimi-code-bench-v2}%
\CensusDot{0}{9}{14}{RadarBlue}{llm-stats-kina}%
\CensusDot{0}{10}{14}{RadarBlue}{llm-stats-labbench2}%
\CensusDot{0}{11}{14}{RadarBlue}{llm-stats-lbpp-v2}%
\CensusDot{0}{12}{14}{RadarBlue}{llm-stats-legal-agent-benchmark}%
\CensusDot{0}{13}{14}{RadarBlue}{llm-stats-lifescibench}%
\CensusDot{0}{14}{14}{RadarBlue}{llm-stats-lingoqa}%
\CensusDot{0}{15}{14}{RadarBlue}{llm-stats-livebench}%
\CensusDot{0}{16}{14}{RadarBlue}{llm-stats-livebench-20241125}%
\CensusDot{0}{17}{14}{RadarBlue}{llm-stats-livecodebench}%
\CensusDot{0}{18}{14}{RadarBlue}{llm-stats-livecodebench-01-09}%
\CensusDot{0}{19}{14}{RadarBlue}{llm-stats-livecodebench-pro}%
\CensusDot{0}{0}{15}{RadarBlue}{llm-stats-livecodebench-v5}%
\CensusDot{0}{1}{15}{RadarBlue}{llm-stats-livecodebench-v5-24-12-25-2}%
\CensusDot{0}{2}{15}{RadarBlue}{llm-stats-livecodebench-v6}%
\CensusDot{0}{3}{15}{RadarBlue}{llm-stats-livemathematicianbench}%
\CensusDot{0}{4}{15}{RadarBlue}{llm-stats-livesports-3k}%
\CensusDot{0}{5}{15}{RadarBlue}{llm-stats-livesqlbench}%
\CensusDot{0}{6}{15}{RadarBlue}{llm-stats-lmarena-text}%
\CensusDot{0}{7}{15}{RadarBlue}{llm-stats-loca-bench-256k}%
\CensusDot{0}{8}{15}{RadarBlue}{llm-stats-longbench-v2}%
\CensusDot{0}{9}{15}{RadarBlue}{llm-stats-longcodebench}%
\CensusDot{0}{10}{15}{RadarBlue}{llm-stats-longfact}%
\CensusDot{0}{11}{15}{RadarBlue}{llm-stats-longfact-concepts}%
\CensusDot{0}{12}{15}{RadarBlue}{llm-stats-longfact-objects}%
\CensusDot{0}{13}{15}{RadarGray}{llm-stats-longtext-bench}%
\CensusDot{0}{14}{15}{RadarBlue}{llm-stats-longvideobench}%
\CensusDot{0}{15}{15}{RadarBlue}{llm-stats-lsat}%
\CensusDot{0}{16}{15}{RadarBlue}{llm-stats-lvbench}%
\CensusDot{0}{17}{15}{RadarBlue}{llm-stats-management-consulting-tasks}%
\CensusDot{0}{18}{15}{RadarBlue}{llm-stats-mask}%
\CensusDot{0}{19}{15}{RadarBlue}{llm-stats-math}%
\CensusDot{0}{0}{16}{RadarBlue}{llm-stats-math-cot}%
\CensusDot{0}{1}{16}{RadarBlue}{llm-stats-math-500}%
\CensusDot{0}{2}{16}{RadarBlue}{llm-stats-matharena-apex}%
\CensusDot{0}{3}{16}{RadarBlue}{llm-stats-mathverse}%
\CensusDot{0}{4}{16}{RadarBlue}{llm-stats-mathverse-mini}%
\CensusDot{0}{5}{16}{RadarBlue}{llm-stats-mathvision}%
\CensusDot{0}{6}{16}{RadarBlue}{llm-stats-mathvista}%
\CensusDot{0}{7}{16}{RadarBlue}{llm-stats-mathvista-mini}%
\CensusDot{0}{8}{16}{RadarBlue}{llm-stats-maverix}%
\CensusDot{0}{9}{16}{RadarBlue}{llm-stats-maxife}%
\CensusDot{0}{10}{16}{RadarBlue}{llm-stats-mbpp}%
\CensusDot{0}{11}{16}{RadarBlue}{llm-stats-mbpp-2}%
\CensusDot{0}{12}{16}{RadarBlue}{llm-stats-mbpp-base-version}%
\CensusDot{0}{13}{16}{RadarBlue}{llm-stats-mbpp-evalplus}%
\CensusDot{0}{14}{16}{RadarBlue}{llm-stats-mbpp-evalplus-base}%
\CensusDot{0}{15}{16}{RadarBlue}{llm-stats-mbpp-pass-1}%
\CensusDot{0}{16}{16}{RadarBlue}{llm-stats-mbpp-plus}%
\CensusDot{0}{17}{16}{RadarBlue}{llm-stats-mcp-atlas}%
\CensusDot{0}{18}{16}{RadarBlue}{llm-stats-mcp-mark}%
\CensusDot{0}{19}{16}{RadarBlue}{llm-stats-mcp-universe}%
\CensusDot{0}{0}{17}{RadarBlue}{llm-stats-measurebench}%
\CensusDot{0}{1}{17}{RadarBlue}{llm-stats-medchembench}%
\CensusDot{0}{2}{17}{RadarBlue}{llm-stats-medxpertqa}%
\CensusDot{0}{3}{17}{RadarBlue}{llm-stats-medxpertqa-mm}%
\CensusDot{0}{4}{17}{RadarBlue}{llm-stats-mega-mlqa}%
\CensusDot{0}{5}{17}{RadarBlue}{llm-stats-mega-tydi-qa}%
\CensusDot{0}{6}{17}{RadarBlue}{llm-stats-mega-udpos}%
\CensusDot{0}{7}{17}{RadarBlue}{llm-stats-mega-xcopa}%
\CensusDot{0}{8}{17}{RadarBlue}{llm-stats-mega-xstorycloze}%
\CensusDot{0}{9}{17}{RadarBlue}{llm-stats-meld}%
\CensusDot{0}{10}{17}{RadarBlue}{llm-stats-meta-internal-coding-bench}%
\CensusDot{0}{11}{17}{RadarBlue}{llm-stats-mewc}%
\CensusDot{0}{12}{17}{RadarBlue}{llm-stats-mgsm}%
\CensusDot{0}{13}{17}{RadarBlue}{llm-stats-miabench}%
\CensusDot{0}{14}{17}{RadarBlue}{llm-stats-mimic-cxr}%
\CensusDot{0}{15}{17}{RadarBlue}{llm-stats-mimo-coding-bench}%
\CensusDot{0}{16}{17}{RadarBlue}{llm-stats-minerva}%
\CensusDot{0}{17}{17}{RadarBlue}{llm-stats-mle-bench}%
\CensusDot{0}{18}{17}{RadarBlue}{llm-stats-mle-bench-lite}%
\CensusDot{0}{19}{17}{RadarBlue}{llm-stats-mls-bench-lite}%
\CensusDot{0}{0}{18}{RadarBlue}{llm-stats-mlvu}%
\CensusDot{0}{1}{18}{RadarBlue}{llm-stats-mlvu-m}%
\CensusDot{0}{2}{18}{RadarBlue}{llm-stats-mm-browsercomp}%
\CensusDot{0}{3}{18}{RadarBlue}{llm-stats-mm-clawbench}%
\CensusDot{0}{4}{18}{RadarBlue}{llm-stats-mm-if-eval}%
\CensusDot{0}{5}{18}{RadarBlue}{llm-stats-mm-mind2web}%
\CensusDot{0}{6}{18}{RadarBlue}{llm-stats-mm-mt-bench}%
\CensusDot{0}{7}{18}{RadarBlue}{llm-stats-mmau}%
\CensusDot{0}{8}{18}{RadarBlue}{llm-stats-mmau-music}%
\CensusDot{0}{9}{18}{RadarBlue}{llm-stats-mmau-sound}%
\CensusDot{0}{10}{18}{RadarBlue}{llm-stats-mmau-speech}%
\CensusDot{0}{11}{18}{RadarBlue}{llm-stats-mmbc}%
\CensusDot{0}{12}{18}{RadarBlue}{llm-stats-mmbench}%
\CensusDot{0}{13}{18}{RadarBlue}{llm-stats-mmbench-v1-1}%
\CensusDot{0}{14}{18}{RadarBlue}{llm-stats-mmbench-video}%
\CensusDot{0}{15}{18}{RadarBlue}{llm-stats-mme}%
\CensusDot{0}{16}{18}{RadarBlue}{llm-stats-mme-realworld}%
\CensusDot{0}{17}{18}{RadarBlue}{llm-stats-mmlongbench-128k}%
\CensusDot{0}{18}{18}{RadarBlue}{llm-stats-mmlongbench-doc}%
\CensusDot{0}{19}{18}{RadarBlue}{llm-stats-mmlu}%
\CensusDot{0}{0}{19}{RadarBlue}{llm-stats-mmlu-cot}%
\CensusDot{0}{1}{19}{RadarBlue}{llm-stats-mmlu-base}%
\CensusDot{0}{2}{19}{RadarBlue}{llm-stats-mmlu-chat}%
\CensusDot{0}{3}{19}{RadarBlue}{llm-stats-mmlu-french}%
\CensusDot{0}{4}{19}{RadarBlue}{llm-stats-mmlu-pro}%
\CensusDot{0}{5}{19}{RadarBlue}{llm-stats-mmlu-prox}%
\CensusDot{0}{6}{19}{RadarBlue}{llm-stats-mmlu-redux}%
\CensusDot{0}{7}{19}{RadarBlue}{llm-stats-mmlu-redux-2-0}%
\CensusDot{0}{8}{19}{RadarBlue}{llm-stats-mmlu-stem}%
\CensusDot{0}{9}{19}{RadarBlue}{llm-stats-mmmlu}%
\CensusDot{0}{10}{19}{RadarBlue}{llm-stats-mmmu}%
\CensusDot{0}{11}{19}{RadarBlue}{llm-stats-mmmu-val}%
\CensusDot{0}{12}{19}{RadarBlue}{llm-stats-mmmu-validation}%
\CensusDot{0}{13}{19}{RadarBlue}{llm-stats-mmmu-pro}%
\CensusDot{0}{14}{19}{RadarBlue}{llm-stats-mmmu-pro-with-tools}%
\CensusDot{0}{15}{19}{RadarBlue}{llm-stats-mmmuval}%
\CensusDot{0}{16}{19}{RadarBlue}{llm-stats-mmsearch}%
\CensusDot{0}{17}{19}{RadarBlue}{llm-stats-mmsearch-plus}%
\CensusDot{0}{18}{19}{RadarBlue}{llm-stats-mmsibench}%
\CensusDot{0}{19}{19}{RadarBlue}{llm-stats-mmstar}%
\CensusDot{0}{0}{20}{RadarBlue}{llm-stats-mmt-bench}%
\CensusDot{0}{1}{20}{RadarBlue}{llm-stats-mmvet}%
\CensusDot{0}{2}{20}{RadarBlue}{llm-stats-mmvetgpt4turbo}%
\CensusDot{0}{3}{20}{RadarBlue}{llm-stats-mmvu}%
\CensusDot{0}{4}{20}{RadarBlue}{llm-stats-mobileminiwob-sr}%
\CensusDot{0}{5}{20}{RadarBlue}{llm-stats-mobileworld}%
\CensusDot{0}{6}{20}{RadarBlue}{llm-stats-motionbench}%
\CensusDot{0}{7}{20}{RadarBlue}{llm-stats-mrcr}%
\CensusDot{0}{8}{20}{RadarBlue}{llm-stats-mrcr-128k-2-needle}%
\CensusDot{0}{9}{20}{RadarBlue}{llm-stats-mrcr-128k-4-needle}%
\CensusDot{0}{10}{20}{RadarBlue}{llm-stats-mrcr-128k-8-needle}%
\CensusDot{0}{11}{20}{RadarBlue}{llm-stats-mrcr-1m}%
\CensusDot{0}{12}{20}{RadarBlue}{llm-stats-mrcr-1m-pointwise}%
\CensusDot{0}{13}{20}{RadarBlue}{llm-stats-mrcr-64k-2-needle}%
\CensusDot{0}{14}{20}{RadarBlue}{llm-stats-mrcr-64k-4-needle}%
\CensusDot{0}{15}{20}{RadarBlue}{llm-stats-mrcr-64k-8-needle}%
\CensusDot{0}{16}{20}{RadarBlue}{llm-stats-mrcr-v2}%
\CensusDot{0}{17}{20}{RadarBlue}{llm-stats-mrcr-v2-8-needle}%
\CensusDot{0}{18}{20}{RadarBlue}{llm-stats-mrcr-v2-8-needle-512k-1m}%
\CensusDot{0}{19}{20}{RadarBlue}{llm-stats-msqa}%
\CensusDot{0}{0}{21}{RadarBlue}{llm-stats-mt-aime-2025}%
\CensusDot{0}{1}{21}{RadarBlue}{llm-stats-mt-bench}%
\CensusDot{0}{2}{21}{RadarBlue}{llm-stats-mtvqa}%
\CensusDot{0}{3}{21}{RadarBlue}{llm-stats-muirbench}%
\CensusDot{0}{4}{21}{RadarBlue}{llm-stats-multi-if}%
\CensusDot{0}{5}{21}{RadarBlue}{llm-stats-multi-swe-bench}%
\CensusDot{0}{6}{21}{RadarBlue}{llm-stats-multichallenge}%
\CensusDot{0}{7}{21}{RadarBlue}{llm-stats-multilf}%
\CensusDot{0}{8}{21}{RadarBlue}{llm-stats-multilingual-mgsm-cot}%
\CensusDot{0}{9}{21}{RadarBlue}{llm-stats-multilingual-mmlu}%
\CensusDot{0}{10}{21}{RadarBlue}{llm-stats-multipl-e}%
\CensusDot{0}{11}{21}{RadarBlue}{llm-stats-multipl-e-humaneval}%
\CensusDot{0}{12}{21}{RadarBlue}{llm-stats-multipl-e-mbpp}%
\CensusDot{0}{13}{21}{RadarBlue}{llm-stats-musiccaps}%
\CensusDot{0}{14}{21}{RadarBlue}{llm-stats-musr}%
\CensusDot{0}{15}{21}{RadarBlue}{llm-stats-mvbench}%
\CensusDot{0}{16}{21}{RadarBlue}{llm-stats-nanogpt}%
\CensusDot{0}{17}{21}{RadarBlue}{llm-stats-natural-questions}%
\CensusDot{0}{18}{21}{RadarBlue}{llm-stats-natural2code}%
\CensusDot{0}{19}{21}{RadarBlue}{llm-stats-nexus}%
\CensusDot{0}{0}{22}{RadarBlue}{llm-stats-nih-multi-needle}%
\CensusDot{0}{1}{22}{RadarBlue}{llm-stats-nl2repo}%
\CensusDot{0}{2}{22}{RadarBlue}{llm-stats-nmos}%
\CensusDot{0}{3}{22}{RadarBlue}{llm-stats-nolima-128k}%
\CensusDot{0}{4}{22}{RadarBlue}{llm-stats-nolima-32k}%
\CensusDot{0}{5}{22}{RadarBlue}{llm-stats-nolima-64k}%
\CensusDot{0}{6}{22}{RadarBlue}{llm-stats-nova-63}%
\CensusDot{0}{7}{22}{RadarBlue}{llm-stats-nq}%
\CensusDot{0}{8}{22}{RadarBlue}{llm-stats-nuscene}%
\CensusDot{0}{9}{22}{RadarBlue}{llm-stats-objectron}%
\CensusDot{0}{10}{22}{RadarBlue}{llm-stats-ocrbench}%
\CensusDot{0}{11}{22}{RadarBlue}{llm-stats-ocrbench-v2}%
\CensusDot{0}{12}{22}{RadarBlue}{llm-stats-ocrbench-v2-en}%
\CensusDot{0}{13}{22}{RadarBlue}{llm-stats-ocrbench-v2-zh}%
\CensusDot{0}{14}{22}{RadarBlue}{llm-stats-octocodingbench}%
\CensusDot{0}{15}{22}{RadarBlue}{llm-stats-odinw}%
\CensusDot{0}{16}{22}{RadarBlue}{llm-stats-officeqa-pro}%
\CensusDot{0}{17}{22}{RadarBlue}{llm-stats-ojbench}%
\CensusDot{0}{18}{22}{RadarBlue}{llm-stats-ojbench-cpp}%
\CensusDot{0}{19}{22}{RadarBlue}{llm-stats-olympiadbench}%
\CensusDot{0}{0}{23}{RadarBlue}{llm-stats-omnibench}%
\CensusDot{0}{1}{23}{RadarBlue}{llm-stats-omnibench-music}%
\CensusDot{0}{2}{23}{RadarBlue}{llm-stats-omnidocbench}%
\CensusDot{0}{3}{23}{RadarBlue}{llm-stats-omnidocbench-1-5}%
\CensusDot{0}{4}{23}{RadarBlue}{llm-stats-omnigaia}%
\CensusDot{0}{5}{23}{RadarBlue}{llm-stats-omnimath}%
\CensusDot{0}{6}{23}{RadarBlue}{llm-stats-omniscience}%
\CensusDot{0}{7}{23}{RadarBlue}{llm-stats-omniscience-non-hallucination-rate}%
\CensusDot{0}{8}{23}{RadarBlue}{llm-stats-onemillion-bench}%
\CensusDot{0}{9}{23}{RadarBlue}{llm-stats-open-rewrite}%
\CensusDot{0}{10}{23}{RadarBlue}{llm-stats-openai-connectors}%
\CensusDot{0}{11}{23}{RadarBlue}{llm-stats-openai-mmlu}%
\CensusDot{0}{12}{23}{RadarBlue}{llm-stats-openai-mrcr-2-needle-128k}%
\CensusDot{0}{13}{23}{RadarBlue}{llm-stats-openai-mrcr-2-needle-1m}%
\CensusDot{0}{14}{23}{RadarBlue}{llm-stats-openai-mrcr-2-needle-256k}%
\CensusDot{0}{15}{23}{RadarBlue}{llm-stats-openai-search-function-calling}%
\CensusDot{0}{16}{23}{RadarBlue}{llm-stats-openbookqa}%
\CensusDot{0}{17}{23}{RadarBlue}{llm-stats-openrca}%
\CensusDot{0}{18}{23}{RadarBlue}{llm-stats-osworld}%
\CensusDot{0}{19}{23}{RadarBlue}{llm-stats-osworld-2-0}%
\CensusDot{0}{0}{24}{RadarBlue}{llm-stats-osworld-extended}%
\CensusDot{0}{1}{24}{RadarBlue}{llm-stats-osworld-g}%
\CensusDot{0}{2}{24}{RadarBlue}{llm-stats-osworld-screenshot-only}%
\CensusDot{0}{3}{24}{RadarBlue}{llm-stats-osworld-verified}%
\CensusDot{0}{4}{24}{RadarBlue}{llm-stats-ovbench}%
\CensusDot{0}{5}{24}{RadarBlue}{llm-stats-ovobench}%
\CensusDot{0}{6}{24}{RadarBlue}{llm-stats-paperbench}%
\CensusDot{0}{7}{24}{RadarBlue}{llm-stats-pathmcqa}%
\CensusDot{0}{8}{24}{RadarBlue}{llm-stats-perceptionbench}%
\CensusDot{0}{9}{24}{RadarBlue}{llm-stats-perceptiontest}%
\CensusDot{0}{10}{24}{RadarBlue}{llm-stats-phibench}%
\CensusDot{0}{11}{24}{RadarBlue}{llm-stats-phybench}%
\CensusDot{0}{12}{24}{RadarBlue}{llm-stats-physicsfinals}%
\CensusDot{0}{13}{24}{RadarBlue}{llm-stats-pinchbench}%
\CensusDot{0}{14}{24}{RadarBlue}{llm-stats-piqa}%
\CensusDot{0}{15}{24}{RadarBlue}{llm-stats-plawbench}%
\CensusDot{0}{16}{24}{RadarBlue}{llm-stats-pmc-vqa}%
\CensusDot{0}{17}{24}{RadarBlue}{llm-stats-pointgrounding}%
\CensusDot{0}{18}{24}{RadarBlue}{llm-stats-polymath}%
\CensusDot{0}{19}{24}{RadarBlue}{llm-stats-polymath-en}%
\CensusDot{0}{0}{25}{RadarBlue}{llm-stats-pope}%
\CensusDot{0}{1}{25}{RadarBlue}{llm-stats-popqa}%
\CensusDot{0}{2}{25}{RadarBlue}{llm-stats-posttrainbench}%
\CensusDot{0}{3}{25}{RadarBlue}{llm-stats-posttrainbench-lite}%
\CensusDot{0}{4}{25}{RadarBlue}{llm-stats-prbench-finance}%
\CensusDot{0}{5}{25}{RadarBlue}{llm-stats-prbench-legal}%
\CensusDot{0}{6}{25}{RadarBlue}{llm-stats-presentbench}%
\CensusDot{0}{7}{25}{RadarBlue}{llm-stats-profbench}%
\CensusDot{0}{8}{25}{RadarBlue}{llm-stats-program-bench}%
\CensusDot{0}{9}{25}{RadarBlue}{llm-stats-protocolqa}%
\CensusDot{0}{10}{25}{RadarBlue}{llm-stats-qasper}%
\CensusDot{0}{11}{25}{RadarBlue}{llm-stats-qmsum}%
\CensusDot{0}{12}{25}{RadarBlue}{llm-stats-qvhighlights}%
\CensusDot{0}{13}{25}{RadarBlue}{llm-stats-qwen-qoder-bench}%
\CensusDot{0}{14}{25}{RadarBlue}{llm-stats-qwen-react-bench}%
\CensusDot{0}{15}{25}{RadarBlue}{llm-stats-qwen-svg}%
\CensusDot{0}{16}{25}{RadarBlue}{llm-stats-qwen-swe-bench}%
\CensusDot{0}{17}{25}{RadarBlue}{llm-stats-qwenclawbench}%
\CensusDot{0}{18}{25}{RadarBlue}{llm-stats-qwenwebbench}%
\CensusDot{0}{19}{25}{RadarBlue}{llm-stats-qwenworldbench}%
\CensusDot{0}{0}{26}{RadarBlue}{llm-stats-realkie-fcc}%
\CensusDot{0}{1}{26}{RadarBlue}{llm-stats-realworldqa}%
\CensusDot{0}{2}{26}{RadarBlue}{llm-stats-recreationbench}%
\CensusDot{0}{3}{26}{RadarBlue}{llm-stats-refcoco-avg}%
\CensusDot{0}{4}{26}{RadarBlue}{llm-stats-refcocog}%
\CensusDot{0}{5}{26}{RadarBlue}{llm-stats-refspatialbench}%
\CensusDot{0}{6}{26}{RadarBlue}{llm-stats-repo-env}%
\CensusDot{0}{7}{26}{RadarBlue}{llm-stats-repobench}%
\CensusDot{0}{8}{26}{RadarBlue}{llm-stats-repoqa}%
\CensusDot{0}{9}{26}{RadarBlue}{llm-stats-researchclawbench}%
\CensusDot{0}{10}{26}{RadarBlue}{llm-stats-robospatialhome}%
\CensusDot{0}{11}{26}{RadarBlue}{llm-stats-robust-if}%
\CensusDot{0}{12}{26}{RadarBlue}{llm-stats-rsi-index}%
\CensusDot{0}{13}{26}{RadarBlue}{llm-stats-ruler}%
\CensusDot{0}{14}{26}{RadarBlue}{llm-stats-ruler-1000k}%
\CensusDot{0}{15}{26}{RadarBlue}{llm-stats-ruler-128k}%
\CensusDot{0}{16}{26}{RadarBlue}{llm-stats-ruler-2048k}%
\CensusDot{0}{17}{26}{RadarBlue}{llm-stats-ruler-512k}%
\CensusDot{0}{18}{26}{RadarBlue}{llm-stats-ruler-64k}%
\CensusDot{0}{19}{26}{RadarBlue}{llm-stats-sat-math}%
\CensusDot{0}{0}{27}{RadarBlue}{llm-stats-scicode}%
\CensusDot{0}{1}{27}{RadarBlue}{llm-stats-scienceqa}%
\CensusDot{0}{2}{27}{RadarBlue}{llm-stats-scienceqa-visual}%
\CensusDot{0}{3}{27}{RadarBlue}{llm-stats-screenspot}%
\CensusDot{0}{4}{27}{RadarBlue}{llm-stats-screenspot-pro}%
\CensusDot{0}{5}{27}{RadarBlue}{llm-stats-seal-0}%
\CensusDot{0}{6}{27}{RadarBlue}{llm-stats-sec-bench-pro}%
\CensusDot{0}{7}{27}{RadarBlue}{llm-stats-seccodebench}%
\CensusDot{0}{8}{27}{RadarBlue}{llm-stats-seedclawbench}%
\CensusDot{0}{9}{27}{RadarBlue}{llm-stats-sifo}%
\CensusDot{0}{10}{27}{RadarBlue}{llm-stats-sifo-multiturn}%
\CensusDot{0}{11}{27}{RadarBlue}{llm-stats-simpleqa}%
\CensusDot{0}{12}{27}{RadarBlue}{llm-stats-simpleqa-verified}%
\CensusDot{0}{13}{27}{RadarBlue}{llm-stats-simplevqa}%
\CensusDot{0}{14}{27}{RadarBlue}{llm-stats-siren-agentdojo-attack-success}%
\CensusDot{0}{15}{27}{RadarBlue}{llm-stats-siren-agentdojo-utility}%
\CensusDot{0}{16}{27}{RadarBlue}{llm-stats-skillsbench}%
\CensusDot{0}{17}{27}{RadarBlue}{llm-stats-slakevqa}%
\CensusDot{0}{18}{27}{RadarBlue}{llm-stats-social-iqa}%
\CensusDot{0}{19}{27}{RadarBlue}{llm-stats-spider}%
\CensusDot{0}{0}{28}{RadarBlue}{llm-stats-spreadsheetbench-2}%
\CensusDot{0}{1}{28}{RadarBlue}{llm-stats-spreadsheetbench-v1}%
\CensusDot{0}{2}{28}{RadarBlue}{llm-stats-squality}%
\CensusDot{0}{3}{28}{RadarBlue}{llm-stats-stem}%
\CensusDot{0}{4}{28}{RadarBlue}{llm-stats-summscreenfd}%
\CensusDot{0}{5}{28}{RadarBlue}{llm-stats-sunrgbd}%
\CensusDot{0}{6}{28}{RadarBlue}{llm-stats-superchem}%
\CensusDot{0}{7}{28}{RadarBlue}{llm-stats-superglue}%
\CensusDot{0}{8}{28}{RadarBlue}{llm-stats-supergpqa}%
\CensusDot{0}{9}{28}{RadarBlue}{llm-stats-surds}%
\CensusDot{0}{10}{28}{RadarBlue}{llm-stats-svg-bench}%
\CensusDot{0}{11}{28}{RadarBlue}{llm-stats-swe-atlas}%
\CensusDot{0}{12}{28}{RadarBlue}{llm-stats-swe-atlas-codebase-qna}%
\CensusDot{0}{13}{28}{RadarBlue}{llm-stats-swe-atlas-test-writing}%
\CensusDot{0}{14}{28}{RadarBlue}{llm-stats-swe-bench-multilingual}%
\CensusDot{0}{15}{28}{RadarBlue}{llm-stats-swe-bench-multimodal}%
\CensusDot{0}{16}{28}{RadarBlue}{llm-stats-swe-bench-pro}%
\CensusDot{0}{17}{28}{RadarBlue}{llm-stats-swe-bench-verified}%
\CensusDot{0}{18}{28}{RadarBlue}{llm-stats-swe-bench-verified-agentic-coding}%
\CensusDot{0}{19}{28}{RadarBlue}{llm-stats-swe-bench-verified-agentless}%
\CensusDot{0}{0}{29}{RadarBlue}{llm-stats-swe-bench-verified-multiple-attempts}%
\CensusDot{0}{1}{29}{RadarBlue}{llm-stats-swe-fficiency}%
\CensusDot{0}{2}{29}{RadarBlue}{llm-stats-swe-lancer}%
\CensusDot{0}{3}{29}{RadarBlue}{llm-stats-swe-lancer-ic-diamond-subset}%
\CensusDot{0}{4}{29}{RadarBlue}{llm-stats-swe-marathon}%
\CensusDot{0}{5}{29}{RadarBlue}{llm-stats-swe-mm}%
\CensusDot{0}{6}{29}{RadarBlue}{llm-stats-swe-perf}%
\CensusDot{0}{7}{29}{RadarBlue}{llm-stats-swe-review}%
\CensusDot{0}{8}{29}{RadarBlue}{llm-stats-swt-bench}%
\CensusDot{0}{9}{29}{RadarBlue}{llm-stats-t2-bench}%
\CensusDot{0}{10}{29}{RadarBlue}{llm-stats-tau-bench}%
\CensusDot{0}{11}{29}{RadarBlue}{llm-stats-tau-bench-airline}%
\CensusDot{0}{12}{29}{RadarBlue}{llm-stats-tau-bench-retail}%
\CensusDot{0}{13}{29}{RadarBlue}{llm-stats-tau2-airline}%
\CensusDot{0}{14}{29}{RadarBlue}{llm-stats-tau2-retail}%
\CensusDot{0}{15}{29}{RadarBlue}{llm-stats-tau2-telecom}%
\CensusDot{0}{16}{29}{RadarBlue}{llm-stats-tau3-airline}%
\CensusDot{0}{17}{29}{RadarBlue}{llm-stats-tau3-banking}%
\CensusDot{0}{18}{29}{RadarBlue}{llm-stats-tau3-bench}%
\CensusDot{0}{19}{29}{RadarBlue}{llm-stats-tau3-retail}%
\CensusDot{0}{0}{30}{RadarBlue}{llm-stats-tau3-telecom}%
\CensusDot{0}{1}{30}{RadarBlue}{llm-stats-tempcompass}%
\CensusDot{0}{2}{30}{RadarBlue}{llm-stats-terminal-bench}%
\CensusDot{0}{3}{30}{RadarBlue}{llm-stats-terminal-bench-2}%
\CensusDot{0}{4}{30}{RadarBlue}{llm-stats-terminal-bench-2-1}%
\CensusDot{0}{5}{30}{RadarBlue}{llm-stats-terminal-bench-3-0}%
\CensusDot{0}{6}{30}{RadarBlue}{llm-stats-terminal-bench-hard}%
\CensusDot{0}{7}{30}{RadarBlue}{llm-stats-terminus}%
\CensusDot{0}{8}{30}{RadarBlue}{llm-stats-textvqa}%
\CensusDot{0}{9}{30}{RadarBlue}{llm-stats-theoremqa}%
\CensusDot{0}{10}{30}{RadarBlue}{llm-stats-tir-bench}%
\CensusDot{0}{11}{30}{RadarBlue}{llm-stats-tldr9-test}%
\CensusDot{0}{12}{30}{RadarBlue}{llm-stats-tomato}%
\CensusDot{0}{13}{30}{RadarBlue}{llm-stats-toolathlon}%
\CensusDot{0}{14}{30}{RadarBlue}{llm-stats-trae-code-gen}%
\CensusDot{0}{15}{30}{RadarBlue}{llm-stats-trae-error-fix}%
\CensusDot{0}{16}{30}{RadarBlue}{llm-stats-translation-en-set1-comet22}%
\CensusDot{0}{17}{30}{RadarBlue}{llm-stats-translation-en-set1-spbleu}%
\CensusDot{0}{18}{30}{RadarBlue}{llm-stats-translation-set1-en-comet22}%
\CensusDot{0}{19}{30}{RadarBlue}{llm-stats-translation-set1-en-spbleu}%
\CensusDot{0}{0}{31}{RadarBlue}{llm-stats-treebench}%
\CensusDot{0}{1}{31}{RadarBlue}{llm-stats-triviaqa}%
\CensusDot{0}{2}{31}{RadarBlue}{llm-stats-truthfulqa}%
\CensusDot{0}{3}{31}{RadarBlue}{llm-stats-tvbench}%
\CensusDot{0}{4}{31}{RadarBlue}{llm-stats-tydiqa}%
\CensusDot{0}{5}{31}{RadarBlue}{llm-stats-uniform-bar-exam}%
\CensusDot{0}{6}{31}{RadarBlue}{llm-stats-usamo-2026}%
\CensusDot{0}{7}{31}{RadarBlue}{llm-stats-usamo25}%
\CensusDot{0}{8}{31}{RadarBlue}{llm-stats-v-star}%
\CensusDot{0}{9}{31}{RadarBlue}{llm-stats-vatex}%
\CensusDot{0}{10}{31}{RadarBlue}{llm-stats-vcr-en-easy}%
\CensusDot{0}{11}{31}{RadarBlue}{llm-stats-vct}%
\CensusDot{0}{12}{31}{RadarBlue}{llm-stats-vending-bench-2}%
\CensusDot{0}{13}{31}{RadarBlue}{llm-stats-vibe}%
\CensusDot{0}{14}{31}{RadarBlue}{llm-stats-vibe-android}%
\CensusDot{0}{15}{31}{RadarBlue}{llm-stats-vibe-backend}%
\CensusDot{0}{16}{31}{RadarBlue}{llm-stats-vibe-eval}%
\CensusDot{0}{17}{31}{RadarBlue}{llm-stats-vibe-ios}%
\CensusDot{0}{18}{31}{RadarBlue}{llm-stats-vibe-pro}%
\CensusDot{0}{19}{31}{RadarBlue}{llm-stats-vibe-simulation}%
\CensusDot{0}{0}{32}{RadarBlue}{llm-stats-vibe-v2}%
\CensusDot{0}{1}{32}{RadarBlue}{llm-stats-vibe-web}%
\CensusDot{0}{2}{32}{RadarBlue}{llm-stats-video-mme}%
\CensusDot{0}{3}{32}{RadarBlue}{llm-stats-video-mme-long-no-subtitles}%
\CensusDot{0}{4}{32}{RadarBlue}{llm-stats-videoholmes}%
\CensusDot{0}{5}{32}{RadarBlue}{llm-stats-videomme-w-o-sub}%
\CensusDot{0}{6}{32}{RadarBlue}{llm-stats-videomme-w-sub}%
\CensusDot{0}{7}{32}{RadarBlue}{llm-stats-videommmu}%
\CensusDot{0}{8}{32}{RadarBlue}{llm-stats-videosimpleqa}%
\CensusDot{0}{9}{32}{RadarBlue}{llm-stats-visfactor}%
\CensusDot{0}{10}{32}{RadarBlue}{llm-stats-vision2web}%
\CensusDot{0}{11}{32}{RadarBlue}{llm-stats-visualwebbench}%
\CensusDot{0}{12}{32}{RadarBlue}{llm-stats-visulogic}%
\CensusDot{0}{13}{32}{RadarBlue}{llm-stats-vita-bench}%
\CensusDot{0}{14}{32}{RadarBlue}{llm-stats-vladbench}%
\CensusDot{0}{15}{32}{RadarBlue}{llm-stats-vlmsarebiased}%
\CensusDot{0}{16}{32}{RadarBlue}{llm-stats-vlmsareblind}%
\CensusDot{0}{17}{32}{RadarBlue}{llm-stats-vocalsound}%
\CensusDot{0}{18}{32}{RadarBlue}{llm-stats-voicebench-avg}%
\CensusDot{0}{19}{32}{RadarBlue}{llm-stats-vqa-rad}%
\CensusDot{0}{0}{33}{RadarBlue}{llm-stats-vqav2}%
\CensusDot{0}{1}{33}{RadarBlue}{llm-stats-vqav2-test}%
\CensusDot{0}{2}{33}{RadarBlue}{llm-stats-vqav2-val}%
\CensusDot{0}{3}{33}{RadarBlue}{llm-stats-we-math}%
\CensusDot{0}{4}{33}{RadarBlue}{llm-stats-web-bench}%
\CensusDot{0}{5}{33}{RadarBlue}{llm-stats-webarena-verified}%
\CensusDot{0}{6}{33}{RadarBlue}{llm-stats-webdev-arena}%
\CensusDot{0}{7}{33}{RadarBlue}{llm-stats-webvoyager}%
\CensusDot{0}{8}{33}{RadarBlue}{llm-stats-widesearch}%
\CensusDot{0}{9}{33}{RadarBlue}{llm-stats-wild-bench}%
\CensusDot{0}{10}{33}{RadarBlue}{llm-stats-wildclawbench}%
\CensusDot{0}{11}{33}{RadarBlue}{llm-stats-winogrande}%
\CensusDot{0}{12}{33}{RadarBlue}{llm-stats-wmdp}%
\CensusDot{0}{13}{33}{RadarBlue}{llm-stats-wmt23}%
\CensusDot{0}{14}{33}{RadarBlue}{llm-stats-wmt24}%
\CensusDot{0}{15}{33}{RadarBlue}{llm-stats-workspace-bench}%
\CensusDot{0}{16}{33}{RadarBlue}{llm-stats-worldbench}%
\CensusDot{0}{17}{33}{RadarBlue}{llm-stats-worldvqa}%
\CensusDot{0}{18}{33}{RadarBlue}{llm-stats-writingbench}%
\CensusDot{0}{19}{33}{RadarBlue}{llm-stats-xdailybench}%
\CensusDot{0}{0}{34}{RadarBlue}{llm-stats-xlsum-english}%
\CensusDot{0}{1}{34}{RadarBlue}{llm-stats-xstest}%
\CensusDot{0}{2}{34}{RadarBlue}{llm-stats-yc-bench}%
\CensusDot{0}{3}{34}{RadarBlue}{llm-stats-zclawbench}%
\CensusDot{0}{4}{34}{RadarBlue}{llm-stats-zebralogic}%
\CensusDot{0}{5}{34}{RadarBlue}{llm-stats-zerobench}%
\CensusDot{0}{6}{34}{RadarBlue}{llm-stats-zerobench-sub}%
\CensusDot{1}{0}{0}{RadarGray}{opencompass-1000-q-bench}%
\CensusDot{1}{1}{0}{RadarGray}{opencompass-1003-s-eval}%
\CensusDot{1}{2}{0}{RadarGray}{opencompass-1017-yue-benchmark}%
\CensusDot{1}{3}{0}{RadarGray}{opencompass-1052-calm}%
\CensusDot{1}{4}{0}{RadarGray}{opencompass-1069-air-bench}%
\CensusDot{1}{5}{0}{RadarGray}{opencompass-1070-olympiadbench}%
\CensusDot{1}{6}{0}{RadarGray}{opencompass-1072-xcodeeval}%
\CensusDot{1}{7}{0}{RadarGray}{opencompass-1073-safetybench}%
\CensusDot{1}{8}{0}{RadarGray}{opencompass-1074-newsbench}%
\CensusDot{1}{9}{0}{RadarGray}{opencompass-1075-alignbench}%
\CensusDot{1}{10}{0}{RadarGray}{opencompass-1076-pca-bench}%
\CensusDot{1}{11}{0}{RadarGray}{opencompass-1077-salad-bench}%
\CensusDot{1}{12}{0}{RadarGray}{opencompass-1078-debugbench}%
\CensusDot{1}{13}{0}{RadarGray}{opencompass-1079-cflue}%
\CensusDot{1}{14}{0}{RadarGray}{opencompass-1080-e-eval}%
\CensusDot{1}{15}{0}{RadarGray}{opencompass-1081-naturalcodebench}%
\CensusDot{1}{16}{0}{RadarGray}{opencompass-1082-studenteval}%
\CensusDot{1}{17}{0}{RadarGray}{opencompass-1083-gaokao-mm}%
\CensusDot{1}{18}{0}{RadarGray}{opencompass-1084-stabletoolbench}%
\CensusDot{1}{19}{0}{RadarGray}{opencompass-1085-infobench}%
\CensusDot{1}{0}{1}{RadarGray}{opencompass-1086-belebele}%
\CensusDot{1}{1}{1}{RadarGray}{opencompass-1087-reveal}%
\CensusDot{1}{2}{1}{RadarGray}{opencompass-1088-uhgeval}%
\CensusDot{1}{3}{1}{RadarGray}{opencompass-1089-mathbench}%
\CensusDot{1}{4}{1}{RadarGray}{opencompass-1091-rolellm}%
\CensusDot{1}{5}{1}{RadarGray}{opencompass-1093-apps}%
\CensusDot{1}{6}{1}{RadarGray}{opencompass-1096-truthfulqa}%
\CensusDot{1}{7}{1}{RadarGray}{opencompass-1097-hellobench}%
\CensusDot{1}{8}{1}{RadarGray}{opencompass-1098-grailqa}%
\CensusDot{1}{9}{1}{RadarGray}{opencompass-1099-mkqa}%
\CensusDot{1}{10}{1}{RadarGray}{opencompass-1100-scienceqa}%
\CensusDot{1}{11}{1}{RadarGray}{opencompass-1102-ms-marco}%
\CensusDot{1}{12}{1}{RadarGray}{opencompass-1103-qasc}%
\CensusDot{1}{13}{1}{RadarGray}{opencompass-1107-strategyqa}%
\CensusDot{1}{14}{1}{RadarGray}{opencompass-1108-hotpotqa}%
\CensusDot{1}{15}{1}{RadarGray}{opencompass-1109-winogrande}%
\CensusDot{1}{16}{1}{RadarGray}{opencompass-1113-svamp}%
\CensusDot{1}{17}{1}{RadarGray}{opencompass-1114-asdiv}%
\CensusDot{1}{18}{1}{RadarGray}{opencompass-1115-mathqa}%
\CensusDot{1}{19}{1}{RadarGray}{opencompass-1116-aqua-rat}%
\CensusDot{1}{0}{2}{RadarGray}{opencompass-1117-naturalproofs}%
\CensusDot{1}{1}{2}{RadarGray}{opencompass-1120-proofnet}%
\CensusDot{1}{2}{2}{RadarGray}{opencompass-1121-halueval}%
\CensusDot{1}{3}{2}{RadarGray}{opencompass-1123-crows-pairs}%
\CensusDot{1}{4}{2}{RadarGray}{opencompass-1124-realtoxicityprompts}%
\CensusDot{1}{5}{2}{RadarGray}{opencompass-1125-mind2web}%
\CensusDot{1}{6}{2}{RadarGray}{opencompass-1128-gorilla}%
\CensusDot{1}{7}{2}{RadarGray}{opencompass-1129-wikisql}%
\CensusDot{1}{8}{2}{RadarGray}{opencompass-1132-tabfact}%
\CensusDot{1}{9}{2}{RadarGray}{opencompass-1133-spider}%
\CensusDot{1}{10}{2}{RadarGray}{opencompass-1134-theoremqa}%
\CensusDot{1}{11}{2}{RadarGray}{opencompass-1135-gpqa}%
\CensusDot{1}{12}{2}{RadarGray}{opencompass-1136-ifeval}%
\CensusDot{1}{13}{2}{RadarGray}{opencompass-1137-cmb}%
\CensusDot{1}{14}{2}{RadarGray}{opencompass-1141-charm}%
\CensusDot{1}{15}{2}{RadarGray}{opencompass-1142-mirage}%
\CensusDot{1}{16}{2}{RadarGray}{opencompass-1145-freb-tqa}%
\CensusDot{1}{17}{2}{RadarGray}{opencompass-1146-bust}%
\CensusDot{1}{18}{2}{RadarGray}{opencompass-1147-m3t}%
\CensusDot{1}{19}{2}{RadarGray}{opencompass-1148-abspyramid}%
\CensusDot{1}{0}{3}{RadarGray}{opencompass-1149-instrusum}%
\CensusDot{1}{1}{3}{RadarGray}{opencompass-1152-sportqa}%
\CensusDot{1}{2}{3}{RadarGray}{opencompass-1155-ada-leval}%
\CensusDot{1}{3}{3}{RadarGray}{opencompass-1164-taskbench}%
\CensusDot{1}{4}{3}{RadarGray}{opencompass-1172-mt-bench-101}%
\CensusDot{1}{5}{3}{RadarGray}{opencompass-1175-mmstar}%
\CensusDot{1}{6}{3}{RadarGray}{opencompass-1178-mathvista}%
\CensusDot{1}{7}{3}{RadarGray}{opencompass-1206-mmbench}%
\CensusDot{1}{8}{3}{RadarGray}{opencompass-1207-mmbench-video}%
\CensusDot{1}{9}{3}{RadarGray}{opencompass-1219-cs-eval}%
\CensusDot{1}{10}{3}{RadarGray}{opencompass-1238-lingoly}%
\CensusDot{1}{11}{3}{RadarGray}{opencompass-1239-cvqa}%
\CensusDot{1}{12}{3}{RadarGray}{opencompass-1241-medcalc-bench}%
\CensusDot{1}{13}{3}{RadarGray}{opencompass-1242-agentboard}%
\CensusDot{1}{14}{3}{RadarGray}{opencompass-1243-embodiedagentinterface}%
\CensusDot{1}{15}{3}{RadarGray}{opencompass-1244-omni-math}%
\CensusDot{1}{16}{3}{RadarGray}{opencompass-1245-kor-bench}%
\CensusDot{1}{17}{3}{RadarGray}{opencompass-1246-livebench}%
\CensusDot{1}{18}{3}{RadarGray}{opencompass-1248-mmmu}%
\CensusDot{1}{19}{3}{RadarGray}{opencompass-1250-collie}%
\CensusDot{1}{0}{4}{RadarGray}{opencompass-1251-planbench}%
\CensusDot{1}{1}{4}{RadarGray}{opencompass-1252-re-bench}%
\CensusDot{1}{2}{4}{RadarGray}{opencompass-1253-bigcodebench}%
\CensusDot{1}{3}{4}{RadarGray}{opencompass-1266-babilong}%
\CensusDot{1}{4}{4}{RadarGray}{opencompass-1267-spreadsheetbench}%
\CensusDot{1}{5}{4}{RadarGray}{opencompass-1268-ctibench}%
\CensusDot{1}{6}{4}{RadarGray}{opencompass-1269-convbench}%
\CensusDot{1}{7}{4}{RadarGray}{opencompass-1270-spider2-v}%
\CensusDot{1}{8}{4}{RadarGray}{opencompass-1272-gsm1k}%
\CensusDot{1}{9}{4}{RadarGray}{opencompass-1273-molpuzzle}%
\CensusDot{1}{10}{4}{RadarGray}{opencompass-1274-imdl-benco}%
\CensusDot{1}{11}{4}{RadarGray}{opencompass-1275-mmlongbench-doc}%
\CensusDot{1}{12}{4}{RadarGray}{opencompass-1276-mmlu-pro}%
\CensusDot{1}{13}{4}{RadarGray}{opencompass-1277-videogui}%
\CensusDot{1}{14}{4}{RadarGray}{opencompass-1278-chronomagic-bench}%
\CensusDot{1}{15}{4}{RadarGray}{opencompass-1279-whodunitbench}%
\CensusDot{1}{16}{4}{RadarGray}{opencompass-1280-ambrosia}%
\CensusDot{1}{17}{4}{RadarGray}{opencompass-1283-p-mmeval}%
\CensusDot{1}{18}{4}{RadarGray}{opencompass-1287-medbench}%
\CensusDot{1}{19}{4}{RadarGray}{opencompass-1317-repliqa}%
\CensusDot{1}{0}{5}{RadarGray}{opencompass-1318-wikicontradict}%
\CensusDot{1}{1}{5}{RadarGray}{opencompass-1319-actionatlas}%
\CensusDot{1}{2}{5}{RadarGray}{opencompass-1320-iac-eval}%
\CensusDot{1}{3}{5}{RadarGray}{opencompass-1321-shoppingmmlu}%
\CensusDot{1}{4}{5}{RadarGray}{opencompass-1322-infibench}%
\CensusDot{1}{5}{5}{RadarGray}{opencompass-1323-scifibench}%
\CensusDot{1}{6}{5}{RadarGray}{opencompass-1324-flub}%
\CensusDot{1}{7}{5}{RadarGray}{opencompass-1325-llm-uncertainty-bench}%
\CensusDot{1}{8}{5}{RadarGray}{opencompass-1326-medjourney}%
\CensusDot{1}{9}{5}{RadarGray}{opencompass-1327-ehrnoteqa}%
\CensusDot{1}{10}{5}{RadarGray}{opencompass-1328-gta}%
\CensusDot{1}{11}{5}{RadarGray}{opencompass-1329-olympicarena}%
\CensusDot{1}{12}{5}{RadarGray}{opencompass-1330-sg-bench}%
\CensusDot{1}{13}{5}{RadarGray}{opencompass-1331-medsafetybench}%
\CensusDot{1}{14}{5}{RadarGray}{opencompass-1332-unibench}%
\CensusDot{1}{15}{5}{RadarGray}{opencompass-1333-redcode}%
\CensusDot{1}{16}{5}{RadarGray}{opencompass-1334-mmdu}%
\CensusDot{1}{17}{5}{RadarGray}{opencompass-1335-ltmbenchmark}%
\CensusDot{1}{18}{5}{RadarGray}{opencompass-1336-compbench}%
\CensusDot{1}{19}{5}{RadarGray}{opencompass-1337-jailtrickbench}%
\CensusDot{1}{0}{6}{RadarGray}{opencompass-1349-cyberseceval}%
\CensusDot{1}{1}{6}{RadarGray}{opencompass-1351-agentharm}%
\CensusDot{1}{2}{6}{RadarGray}{opencompass-1355-hallusionbench}%
\CensusDot{1}{3}{6}{RadarGray}{opencompass-1356-mm-vet}%
\CensusDot{1}{4}{6}{RadarGray}{opencompass-1357-mme}%
\CensusDot{1}{5}{6}{RadarGray}{opencompass-1358-video-mme}%
\CensusDot{1}{6}{6}{RadarGray}{opencompass-1359-seed-bench}%
\CensusDot{1}{7}{6}{RadarGray}{opencompass-1360-llava-bench}%
\CensusDot{1}{8}{6}{RadarGray}{opencompass-1361-realworldqa}%
\CensusDot{1}{9}{6}{RadarGray}{opencompass-1362-pope}%
\CensusDot{1}{10}{6}{RadarGray}{opencompass-1363-seed-bench-2}%
\CensusDot{1}{11}{6}{RadarGray}{opencompass-1364-mmt-bench}%
\CensusDot{1}{12}{6}{RadarGray}{opencompass-1365-blink}%
\CensusDot{1}{13}{6}{RadarGray}{opencompass-1367-a-okvqa}%
\CensusDot{1}{14}{6}{RadarGray}{opencompass-1370-mathvision}%
\CensusDot{1}{15}{6}{RadarGray}{opencompass-1371-mathverse}%
\CensusDot{1}{16}{6}{RadarGray}{opencompass-1374-dynamath}%
\CensusDot{1}{17}{6}{RadarGray}{opencompass-1375-vbench}%
\CensusDot{1}{18}{6}{RadarGray}{opencompass-1376-genai-bench}%
\CensusDot{1}{19}{6}{RadarGray}{opencompass-1395-seed-bench-2-plus}%
\CensusDot{1}{0}{7}{RadarGray}{opencompass-1396-av-odyssey-bench}%
\CensusDot{1}{1}{7}{RadarGray}{opencompass-1397-livemathbench}%
\CensusDot{1}{2}{7}{RadarGray}{opencompass-1413-livecodebench}%
\CensusDot{1}{3}{7}{RadarGray}{opencompass-1451-mme-realworld}%
\CensusDot{1}{4}{7}{RadarGray}{opencompass-1453-mmiu}%
\CensusDot{1}{5}{7}{RadarGray}{opencompass-1500-crpe}%
\CensusDot{1}{6}{7}{RadarGray}{opencompass-1502-mtvqa}%
\CensusDot{1}{7}{7}{RadarGray}{opencompass-1509-mvbench}%
\CensusDot{1}{8}{7}{RadarGray}{opencompass-1510-longvideobench}%
\CensusDot{1}{9}{7}{RadarGray}{opencompass-1512-mlvu}%
\CensusDot{1}{10}{7}{RadarGray}{opencompass-1513-cg-bench}%
\CensusDot{1}{11}{7}{RadarGray}{opencompass-1518-minictx}%
\CensusDot{1}{12}{7}{RadarGray}{opencompass-1523-mmie}%
\CensusDot{1}{13}{7}{RadarGray}{opencompass-1524-rm-bench}%
\CensusDot{1}{14}{7}{RadarGray}{opencompass-1532-zerobench}%
\CensusDot{1}{15}{7}{RadarGray}{opencompass-1533-nutritionqa}%
\CensusDot{1}{16}{7}{RadarGray}{opencompass-1534-chase-code}%
\CensusDot{1}{17}{7}{RadarGray}{opencompass-1537-mvl-sib}%
\CensusDot{1}{18}{7}{RadarGray}{opencompass-1538-supergpqa}%
\CensusDot{1}{19}{7}{RadarGray}{opencompass-1539-mm-rlhf}%
\CensusDot{1}{0}{8}{RadarGray}{opencompass-1541-vlm2-bench}%
\CensusDot{1}{1}{8}{RadarGray}{opencompass-1542-structflowbench}%
\CensusDot{1}{2}{8}{RadarGray}{opencompass-1543-kitab-bench}%
\CensusDot{1}{3}{8}{RadarGray}{opencompass-1546-codecriticbench}%
\CensusDot{1}{4}{8}{RadarGray}{opencompass-1547-mmir}%
\CensusDot{1}{5}{8}{RadarGray}{opencompass-1548-medhallu}%
\CensusDot{1}{6}{8}{RadarGray}{opencompass-1552-ceb}%
\CensusDot{1}{7}{8}{RadarGray}{opencompass-1553-omnialign-v}%
\CensusDot{1}{8}{8}{RadarGray}{opencompass-1554-jl1-cd}%
\CensusDot{1}{9}{8}{RadarGray}{opencompass-1555-wildbench}%
\CensusDot{1}{10}{8}{RadarGray}{opencompass-1557-airbench-2024}%
\CensusDot{1}{11}{8}{RadarGray}{opencompass-1558-mm-alignbench}%
\CensusDot{1}{12}{8}{RadarGray}{opencompass-1562-benchmax}%
\CensusDot{1}{13}{8}{RadarGray}{opencompass-1564-embodiedbench}%
\CensusDot{1}{14}{8}{RadarGray}{opencompass-1565-mme-cot}%
\CensusDot{1}{15}{8}{RadarGray}{opencompass-1566-mm-iq}%
\CensusDot{1}{16}{8}{RadarGray}{opencompass-1571-bright}%
\CensusDot{1}{17}{8}{RadarGray}{opencompass-1572-loki}%
\CensusDot{1}{18}{8}{RadarGray}{opencompass-1574-text2world}%
\CensusDot{1}{19}{8}{RadarGray}{opencompass-1576-physreason}%
\CensusDot{1}{0}{9}{RadarGray}{opencompass-1578-mmke-bench}%
\CensusDot{1}{1}{9}{RadarGray}{opencompass-1579-postersum}%
\CensusDot{1}{2}{9}{RadarGray}{opencompass-1580-a-bench}%
\CensusDot{1}{3}{9}{RadarGray}{opencompass-1581-holobench}%
\CensusDot{1}{4}{9}{RadarGray}{opencompass-1582-codemmlu}%
\CensusDot{1}{5}{9}{RadarGray}{opencompass-1583-judgebench}%
\CensusDot{1}{6}{9}{RadarGray}{opencompass-1584-mmsearch}%
\CensusDot{1}{7}{9}{RadarGray}{opencompass-1585-mmad}%
\CensusDot{1}{8}{9}{RadarGray}{opencompass-1586-mrag-bench}%
\CensusDot{1}{9}{9}{RadarGray}{opencompass-1587-mr-gsm8k}%
\CensusDot{1}{10}{9}{RadarGray}{opencompass-1599-gaia}%
\CensusDot{1}{11}{9}{RadarGray}{opencompass-1604-egonormia}%
\CensusDot{1}{12}{9}{RadarGray}{opencompass-1605-deepfake-eval-2024}%
\CensusDot{1}{13}{9}{RadarGray}{opencompass-1606-mcitebench}%
\CensusDot{1}{14}{9}{RadarGray}{opencompass-1607-toolret}%
\CensusDot{1}{15}{9}{RadarGray}{opencompass-1608-swiltra-bench}%
\CensusDot{1}{16}{9}{RadarGray}{opencompass-1609-mask}%
\CensusDot{1}{17}{9}{RadarGray}{opencompass-1617-ifir}%
\CensusDot{1}{18}{9}{RadarGray}{opencompass-1618-fedmabench}%
\CensusDot{1}{19}{9}{RadarGray}{opencompass-1621-processbench}%
\CensusDot{1}{0}{10}{RadarGray}{opencompass-1622-coral}%
\CensusDot{1}{1}{10}{RadarGray}{opencompass-1623-mj-bench}%
\CensusDot{1}{2}{10}{RadarGray}{opencompass-1624-codeelo}%
\CensusDot{1}{3}{10}{RadarGray}{opencompass-1625-ubuntu-osworld}%
\CensusDot{1}{4}{10}{RadarGray}{opencompass-1626-crag}%
\CensusDot{1}{5}{10}{RadarGray}{opencompass-1632-probench}%
\CensusDot{1}{6}{10}{RadarGray}{opencompass-1633-projudge}%
\CensusDot{1}{7}{10}{RadarGray}{opencompass-1634-visualsimpleqa}%
\CensusDot{1}{8}{10}{RadarGray}{opencompass-1635-knowlogic}%
\CensusDot{1}{9}{10}{RadarGray}{opencompass-1636-urbanvideo-bench}%
\CensusDot{1}{10}{10}{RadarGray}{opencompass-1640-emma}%
\CensusDot{1}{11}{10}{RadarGray}{opencompass-1641-medagents-bench}%
\CensusDot{1}{12}{10}{RadarGray}{opencompass-1643-creation-mmbench}%
\CensusDot{1}{13}{10}{RadarGray}{opencompass-1645-mastermindeval}%
\CensusDot{1}{14}{10}{RadarGray}{opencompass-1648-dme}%
\CensusDot{1}{15}{10}{RadarGray}{opencompass-1654-v-star}%
\CensusDot{1}{16}{10}{RadarGray}{opencompass-1656-rfuav}%
\CensusDot{1}{17}{10}{RadarGray}{opencompass-1658-milic-eval}%
\CensusDot{1}{18}{10}{RadarGray}{opencompass-1667-microvqa}%
\CensusDot{1}{19}{10}{RadarGray}{opencompass-1668-timetravel}%
\CensusDot{1}{0}{11}{RadarGray}{opencompass-1670-indicmmlu-pro}%
\CensusDot{1}{1}{11}{RadarGray}{opencompass-1671-forensics-bench}%
\CensusDot{1}{2}{11}{RadarGray}{opencompass-1675-pokerbench}%
\CensusDot{1}{3}{11}{RadarGray}{opencompass-1678-rsmmvp}%
\CensusDot{1}{4}{11}{RadarGray}{opencompass-1679-contextualjudgebench}%
\CensusDot{1}{5}{11}{RadarGray}{opencompass-1680-bigobench}%
\CensusDot{1}{6}{11}{RadarGray}{opencompass-1681-prmbench-preview}%
\CensusDot{1}{7}{11}{RadarGray}{opencompass-1682-motionbench}%
\CensusDot{1}{8}{11}{RadarGray}{opencompass-1694-maritimebench}%
\CensusDot{1}{9}{11}{RadarGray}{opencompass-1701-writingbench}%
\CensusDot{1}{10}{11}{RadarGray}{opencompass-1704-mono2stereo}%
\CensusDot{1}{11}{11}{RadarGray}{opencompass-1705-olymmath}%
\CensusDot{1}{12}{11}{RadarGray}{opencompass-1706-koffvqa}%
\CensusDot{1}{13}{11}{RadarGray}{opencompass-1707-rxrx3-core}%
\CensusDot{1}{14}{11}{RadarGray}{opencompass-1712-toolhop}%
\CensusDot{1}{15}{11}{RadarGray}{opencompass-1717-vilbench}%
\CensusDot{1}{16}{11}{RadarGray}{opencompass-1730-paperbench}%
\CensusDot{1}{17}{11}{RadarGray}{opencompass-1731-dove}%
\CensusDot{1}{18}{11}{RadarGray}{opencompass-1732-scam}%
\CensusDot{1}{19}{11}{RadarGray}{opencompass-1733-feabench}%
\CensusDot{1}{0}{12}{RadarGray}{opencompass-1734-thai-local-benchmark}%
\CensusDot{1}{1}{12}{RadarGray}{opencompass-1735-worldscore}%
\CensusDot{1}{2}{12}{RadarGray}{opencompass-1736-mmtb}%
\CensusDot{1}{3}{12}{RadarGray}{opencompass-1737-fortisavqa}%
\CensusDot{1}{4}{12}{RadarGray}{opencompass-1738-rulistening}%
\CensusDot{1}{5}{12}{RadarGray}{opencompass-1739-crosswordbench}%
\CensusDot{1}{6}{12}{RadarGray}{opencompass-1740-gpt-imgeval}%
\CensusDot{1}{7}{12}{RadarGray}{opencompass-1742-u-niah}%
\CensusDot{1}{8}{12}{RadarGray}{opencompass-1744-stylerec}%
\CensusDot{1}{9}{12}{RadarGray}{opencompass-1748-visualpuzzles}%
\CensusDot{1}{10}{12}{RadarGray}{opencompass-1751-multiloko}%
\CensusDot{1}{11}{12}{RadarGray}{opencompass-1752-llm-srbench}%
\CensusDot{1}{12}{12}{RadarGray}{opencompass-1753-agmmu}%
\CensusDot{1}{13}{12}{RadarGray}{opencompass-1754-openturingbench}%
\CensusDot{1}{14}{12}{RadarGray}{opencompass-1755-real}%
\CensusDot{1}{15}{12}{RadarGray}{opencompass-1772-s1-bench}%
\CensusDot{1}{16}{12}{RadarGray}{opencompass-1777-colorbench}%
\CensusDot{1}{17}{12}{RadarGray}{opencompass-1778-agentrewardbench}%
\CensusDot{1}{18}{12}{RadarGray}{opencompass-1779-mlrc-bench}%
\CensusDot{1}{19}{12}{RadarGray}{opencompass-1780-c-faith}%
\CensusDot{1}{0}{13}{RadarGray}{opencompass-1781-mieb}%
\CensusDot{1}{1}{13}{RadarGray}{opencompass-1782-hypobench}%
\CensusDot{1}{2}{13}{RadarGray}{opencompass-1783-nppc}%
\CensusDot{1}{3}{13}{RadarGray}{opencompass-1784-xverify}%
\CensusDot{1}{4}{13}{RadarGray}{opencompass-1786-hypoeval}%
\CensusDot{1}{5}{13}{RadarGray}{opencompass-1787-livelongbench}%
\CensusDot{1}{6}{13}{RadarGray}{opencompass-1801-omnigirl}%
\CensusDot{1}{7}{13}{RadarGray}{opencompass-1829-mms-vpr}%
\CensusDot{1}{8}{13}{RadarGray}{opencompass-1832-medbrowsecomp}%
\CensusDot{1}{9}{13}{RadarGray}{opencompass-1833-csts}%
\CensusDot{1}{10}{13}{RadarGray}{opencompass-1834-clever}%
\CensusDot{1}{11}{13}{RadarGray}{opencompass-1835-audiojailbreak}%
\CensusDot{1}{12}{13}{RadarGray}{opencompass-1836-pashtoocr}%
\CensusDot{1}{13}{13}{RadarGray}{opencompass-1838-llm-babybench}%
\CensusDot{1}{14}{13}{RadarGray}{opencompass-1839-tiny-qa-benchmark-pp}%
\CensusDot{1}{15}{13}{RadarGray}{opencompass-1840-iqbench}%
\CensusDot{1}{16}{13}{RadarGray}{opencompass-1842-hardmath2}%
\CensusDot{1}{17}{13}{RadarGray}{opencompass-1843-ewmbench}%
\CensusDot{1}{18}{13}{RadarGray}{opencompass-1844-miracl-vision}%
\CensusDot{1}{19}{13}{RadarGray}{opencompass-1845-stark-10k}%
\CensusDot{1}{0}{14}{RadarGray}{opencompass-1846-tglg}%
\CensusDot{1}{1}{14}{RadarGray}{opencompass-1847-massive-steps}%
\CensusDot{1}{2}{14}{RadarGray}{opencompass-1848-mavos-dd}%
\CensusDot{1}{3}{14}{RadarGray}{opencompass-1849-cleanpatrick}%
\CensusDot{1}{4}{14}{RadarGray}{opencompass-1851-mmlongbench}%
\CensusDot{1}{5}{14}{RadarGray}{opencompass-1853-minilongbench}%
\CensusDot{1}{6}{14}{RadarGray}{opencompass-1855-transbench}%
\CensusDot{1}{7}{14}{RadarGray}{opencompass-1865-gitgoodbench}%
\CensusDot{1}{8}{14}{RadarGray}{opencompass-1874-medarabiq}%
\CensusDot{1}{9}{14}{RadarGray}{opencompass-1875-vibe}%
\CensusDot{1}{10}{14}{RadarGray}{opencompass-1876-aneumo}%
\CensusDot{1}{11}{14}{RadarGray}{opencompass-1891-medxpertqa}%
\CensusDot{1}{12}{14}{RadarGray}{opencompass-1892-er-reason}%
\CensusDot{1}{13}{14}{RadarGray}{opencompass-1893-medal}%
\CensusDot{1}{14}{14}{RadarGray}{opencompass-1900-videoreasonbench}%
\CensusDot{1}{15}{14}{RadarGray}{opencompass-1905-mvpbench}%
\CensusDot{1}{16}{14}{RadarGray}{opencompass-1907-cfinbench}%
\CensusDot{1}{17}{14}{RadarGray}{opencompass-1908-audiotrust}%
\CensusDot{1}{18}{14}{RadarGray}{opencompass-1910-mmar}%
\CensusDot{1}{19}{14}{RadarGray}{opencompass-1915-rewardbench}%
\CensusDot{1}{0}{15}{RadarGray}{opencompass-1917-orak}%
\CensusDot{1}{1}{15}{RadarGray}{opencompass-1918-rdb2g-bench}%
\CensusDot{1}{2}{15}{RadarGray}{opencompass-1920-medbookvqa}%
\CensusDot{1}{3}{15}{RadarGray}{opencompass-1922-mmsi-bench}%
\CensusDot{1}{4}{15}{RadarGray}{opencompass-1924-lamp-qa}%
\CensusDot{1}{5}{15}{RadarGray}{opencompass-1926-videomathqa}%
\CensusDot{1}{6}{15}{RadarGray}{opencompass-1927-gmaimmbench}%
\CensusDot{1}{7}{15}{RadarGray}{opencompass-1929-mtcmb}%
\CensusDot{1}{8}{15}{RadarGray}{opencompass-1938-worldgenbench}%
\CensusDot{1}{9}{15}{RadarGray}{opencompass-1942-omnidocbench}%
\CensusDot{1}{10}{15}{RadarGray}{opencompass-1943-combibench}%
\CensusDot{1}{11}{15}{RadarGray}{opencompass-1944-loopnav}%
\CensusDot{1}{12}{15}{RadarGray}{opencompass-1954-climateviz}%
\CensusDot{1}{13}{15}{RadarGray}{opencompass-1959-personalens}%
\CensusDot{1}{14}{15}{RadarGray}{opencompass-1961-sfe}%
\CensusDot{1}{15}{15}{RadarGray}{opencompass-1966-amsbench}%
\CensusDot{1}{16}{15}{RadarGray}{opencompass-1969-ale-bench}%
\CensusDot{1}{17}{15}{RadarGray}{opencompass-1970-editinspector}%
\CensusDot{1}{18}{15}{RadarGray}{opencompass-1971-dycodeeval}%
\CensusDot{1}{19}{15}{RadarGray}{opencompass-1972-cpret}%
\CensusDot{1}{0}{16}{RadarGray}{opencompass-1975-causalvqa}%
\CensusDot{1}{1}{16}{RadarGray}{opencompass-1976-intphys2}%
\CensusDot{1}{2}{16}{RadarGray}{opencompass-1979-openunlearning}%
\CensusDot{1}{3}{16}{RadarGray}{opencompass-1980-falsereject}%
\CensusDot{1}{4}{16}{RadarGray}{opencompass-1981-omnibench}%
\CensusDot{1}{5}{16}{RadarGray}{opencompass-1982-webui-bench}%
\CensusDot{1}{6}{16}{RadarGray}{opencompass-1983-bytemorph}%
\CensusDot{1}{7}{16}{RadarGray}{opencompass-1984-swe-bench-live}%
\CensusDot{1}{8}{16}{RadarGray}{opencompass-1985-cvdp}%
\CensusDot{1}{9}{16}{RadarGray}{opencompass-1986-sec-bench}%
\CensusDot{1}{10}{16}{RadarGray}{opencompass-1987-deepresearchbench}%
\CensusDot{1}{11}{16}{RadarGray}{opencompass-1989-swe-factory}%
\CensusDot{1}{12}{16}{RadarGray}{opencompass-1990-opt-bench}%
\CensusDot{1}{13}{16}{RadarGray}{opencompass-1991-assetopsbench}%
\CensusDot{1}{14}{16}{RadarGray}{opencompass-1992-htfllib}%
\CensusDot{1}{15}{16}{RadarGray}{opencompass-1993-vistorybench}%
\CensusDot{1}{16}{16}{RadarGray}{opencompass-1995-airtbench}%
\CensusDot{1}{17}{16}{RadarGray}{opencompass-1999-morse-500}%
\CensusDot{1}{18}{16}{RadarGray}{opencompass-2005-oss-bench}%
\CensusDot{1}{19}{16}{RadarGray}{opencompass-2014-utboost}%
\CensusDot{1}{0}{17}{RadarGray}{opencompass-2025-tableeval}%
\CensusDot{1}{1}{17}{RadarGray}{opencompass-2027-groundingsuite}%
\CensusDot{1}{2}{17}{RadarGray}{opencompass-2029-smmile}%
\CensusDot{1}{3}{17}{RadarGray}{opencompass-2030-dabstep}%
\CensusDot{1}{4}{17}{RadarGray}{opencompass-2031-herb}%
\CensusDot{1}{5}{17}{RadarGray}{opencompass-2032-dice-bench}%
\CensusDot{1}{6}{17}{RadarGray}{opencompass-2033-rexbench}%
\CensusDot{1}{7}{17}{RadarGray}{opencompass-2034-mteb}%
\CensusDot{1}{8}{17}{RadarGray}{opencompass-2035-gym4real}%
\CensusDot{1}{9}{17}{RadarGray}{opencompass-2039-translaw}%
\CensusDot{1}{10}{17}{RadarGray}{opencompass-2045-dragon}%
\CensusDot{1}{11}{17}{RadarGray}{opencompass-2046-crew-wildfire}%
\CensusDot{1}{12}{17}{RadarGray}{opencompass-2047-llmthinkbench}%
\CensusDot{1}{13}{17}{RadarGray}{opencompass-2048-risebench}%
\CensusDot{1}{14}{17}{RadarGray}{opencompass-2050-thunder}%
\CensusDot{1}{15}{17}{RadarGray}{opencompass-2052-artifactsbench}%
\CensusDot{1}{16}{17}{RadarGray}{opencompass-2061-agenthazard}%
\CensusDot{1}{17}{17}{RadarGray}{opencompass-2063-visco}%
\CensusDot{1}{18}{17}{RadarGray}{opencompass-2067-longvale}%
\CensusDot{1}{19}{17}{RadarGray}{opencompass-2073-j1-bench}%
\CensusDot{1}{0}{18}{RadarGray}{opencompass-2075-arena-hard-auto}%
\CensusDot{1}{1}{18}{RadarGray}{opencompass-2076-k-sort-arena}%
\CensusDot{1}{2}{18}{RadarGray}{opencompass-2077-itbench}%
\CensusDot{1}{3}{18}{RadarGray}{opencompass-2078-autoadvexbench}%
\CensusDot{1}{4}{18}{RadarGray}{opencompass-2079-mer-unibench}%
\CensusDot{1}{5}{18}{RadarGray}{opencompass-2080-phygenbench}%
\CensusDot{1}{6}{18}{RadarGray}{opencompass-2081-general-bench}%
\CensusDot{1}{7}{18}{RadarGray}{opencompass-2082-structtokenbench}%
\CensusDot{1}{8}{18}{RadarGray}{opencompass-2083-saebench}%
\CensusDot{1}{9}{18}{RadarGray}{opencompass-2084-axbench}%
\CensusDot{1}{10}{18}{RadarGray}{opencompass-2085-mib}%
\CensusDot{1}{11}{18}{RadarGray}{opencompass-2086-or-bench}%
\CensusDot{1}{12}{18}{RadarGray}{opencompass-2087-perteval-scfm}%
\CensusDot{1}{13}{18}{RadarGray}{opencompass-2088-lmact}%
\CensusDot{1}{14}{18}{RadarGray}{opencompass-2089-lara}%
\CensusDot{1}{15}{18}{RadarGray}{opencompass-2091-is-bench}%
\CensusDot{1}{16}{18}{RadarGray}{opencompass-2092-spatial457}%
\CensusDot{1}{17}{18}{RadarGray}{opencompass-2126-kmmlu-redux}%
\CensusDot{1}{18}{18}{RadarGray}{opencompass-2127-clembench}%
\CensusDot{1}{19}{18}{RadarGray}{opencompass-2128-lm-evaluation-harness}%
\CensusDot{1}{0}{19}{RadarGray}{opencompass-2129-langnavbench}%
\CensusDot{1}{1}{19}{RadarGray}{opencompass-2130-job-complex}%
\CensusDot{1}{2}{19}{RadarGray}{opencompass-2142-omnimmi}%
\CensusDot{1}{3}{19}{RadarGray}{opencompass-2151-interndata-a1}%
\CensusDot{1}{4}{19}{RadarGray}{opencompass-2152-interndata-n1}%
\CensusDot{1}{5}{19}{RadarGray}{opencompass-2153-interndata-m1}%
\CensusDot{1}{6}{19}{RadarGray}{opencompass-2154-motionmillion}%
\CensusDot{1}{7}{19}{RadarGray}{opencompass-2217-smartbench}%
\CensusDot{1}{8}{19}{RadarGray}{opencompass-2233-gsm8k-v}%
\CensusDot{1}{9}{19}{RadarGray}{opencompass-2325-sgi-bench}%
\CensusDot{1}{10}{19}{RadarGray}{opencompass-2348-shell}%
\CensusDot{1}{11}{19}{RadarGray}{opencompass-2350-picabench}%
\CensusDot{1}{12}{19}{RadarGray}{opencompass-2370-argusinspection}%
\CensusDot{1}{13}{19}{RadarGray}{opencompass-2371-rigorousbench}%
\CensusDot{1}{14}{19}{RadarGray}{opencompass-2383-vrbench}%
\CensusDot{1}{15}{19}{RadarGray}{opencompass-2388-vknowu}%
\CensusDot{1}{16}{19}{RadarGray}{opencompass-2391-threat-signature-eval}%
\CensusDot{1}{17}{19}{RadarGray}{opencompass-2396-aidabench}%
\CensusDot{1}{18}{19}{RadarGray}{opencompass-2417-medhalltune}%
\CensusDot{1}{19}{19}{RadarGray}{opencompass-2422-lens}%
\CensusDot{1}{0}{20}{RadarGray}{opencompass-2445-wildclawbench}%
\CensusDot{1}{1}{20}{RadarGray}{opencompass-2452-aecbench}%
\CensusDot{1}{2}{20}{RadarGray}{opencompass-2571-elbench}%
\CensusDot{1}{3}{20}{RadarGray}{opencompass-2574-gauge}%
\CensusDot{1}{4}{20}{RadarGray}{opencompass-496-c-eval}%
\CensusDot{1}{5}{20}{RadarGray}{opencompass-497-agieval}%
\CensusDot{1}{6}{20}{RadarGray}{opencompass-498-mmlu}%
\CensusDot{1}{7}{20}{RadarGray}{opencompass-499-cmmlu}%
\CensusDot{1}{8}{20}{RadarGray}{opencompass-500-gaokao-bench}%
\CensusDot{1}{9}{20}{RadarGray}{opencompass-502-arc-c}%
\CensusDot{1}{10}{20}{RadarGray}{opencompass-503-arc-e}%
\CensusDot{1}{11}{20}{RadarGray}{opencompass-504-wic}%
\CensusDot{1}{12}{20}{RadarGray}{opencompass-505-chid}%
\CensusDot{1}{13}{20}{RadarGray}{opencompass-506-afqmc}%
\CensusDot{1}{14}{20}{RadarGray}{opencompass-507-wsc}%
\CensusDot{1}{15}{20}{RadarGray}{opencompass-508-tydiqa}%
\CensusDot{1}{16}{20}{RadarGray}{opencompass-509-flores}%
\CensusDot{1}{17}{20}{RadarGray}{opencompass-510-boolq}%
\CensusDot{1}{18}{20}{RadarGray}{opencompass-511-commonsenseqa}%
\CensusDot{1}{19}{20}{RadarGray}{opencompass-512-triviaqa}%
\CensusDot{1}{0}{21}{RadarGray}{opencompass-513-nq}%
\CensusDot{1}{1}{21}{RadarGray}{opencompass-514-c3}%
\CensusDot{1}{2}{21}{RadarGray}{opencompass-516-race-high}%
\CensusDot{1}{3}{21}{RadarGray}{opencompass-517-race-middle}%
\CensusDot{1}{4}{21}{RadarGray}{opencompass-518-openbookqa}%
\CensusDot{1}{5}{21}{RadarGray}{opencompass-519-csl}%
\CensusDot{1}{6}{21}{RadarGray}{opencompass-520-lcsts}%
\CensusDot{1}{7}{21}{RadarGray}{opencompass-521-xsum}%
\CensusDot{1}{8}{21}{RadarGray}{opencompass-522-eprstmt}%
\CensusDot{1}{9}{21}{RadarGray}{opencompass-523-lambada}%
\CensusDot{1}{10}{21}{RadarGray}{opencompass-524-cmnli}%
\CensusDot{1}{11}{21}{RadarGray}{opencompass-525-ocnli}%
\CensusDot{1}{12}{21}{RadarGray}{opencompass-526-ax-b}%
\CensusDot{1}{13}{21}{RadarGray}{opencompass-527-ax-g}%
\CensusDot{1}{14}{21}{RadarGray}{opencompass-528-rte}%
\CensusDot{1}{15}{21}{RadarGray}{opencompass-529-copa}%
\CensusDot{1}{16}{21}{RadarGray}{opencompass-530-record}%
\CensusDot{1}{17}{21}{RadarGray}{opencompass-531-hellaswag}%
\CensusDot{1}{18}{21}{RadarGray}{opencompass-532-piqa}%
\CensusDot{1}{19}{21}{RadarGray}{opencompass-533-siqa}%
\CensusDot{1}{0}{22}{RadarGray}{opencompass-534-math}%
\CensusDot{1}{1}{22}{RadarGray}{opencompass-535-gsm8k}%
\CensusDot{1}{2}{22}{RadarGray}{opencompass-536-drop}%
\CensusDot{1}{3}{22}{RadarGray}{opencompass-537-humaneval}%
\CensusDot{1}{4}{22}{RadarGray}{opencompass-538-mbpp}%
\CensusDot{1}{5}{22}{RadarGray}{opencompass-539-bbh}%
\CensusDot{1}{6}{22}{RadarGray}{opencompass-540-t-eval}%
\CensusDot{1}{7}{22}{RadarGray}{opencompass-541-l-eval}%
\CensusDot{1}{8}{22}{RadarGray}{opencompass-542-longbench}%
\CensusDot{1}{9}{22}{RadarGray}{opencompass-543-humaneval-x}%
\CensusDot{1}{10}{22}{RadarGray}{opencompass-544-ds-1000}%
\CensusDot{1}{11}{22}{RadarGray}{opencompass-557-ocrbench}%
\CensusDot{1}{12}{22}{RadarGray}{opencompass-564-lv-eval}%
\CensusDot{1}{13}{22}{RadarGray}{opencompass-568-criticbench}%
\CensusDot{1}{14}{22}{RadarGray}{opencompass-631-openfindata}%
\CensusDot{1}{15}{22}{RadarGray}{opencompass-692-chembench}%
\CensusDot{1}{16}{22}{RadarGray}{opencompass-895-fin-eva}%
\CensusDot{1}{17}{22}{RadarGray}{opencompass-924-cs-bench}%
\CensusDot{1}{18}{22}{RadarGray}{opencompass-930-mr-ben-meta-reasoning-benchmark}%
\CensusDot{1}{19}{22}{RadarGray}{opencompass-945-flames}%
\CensusDot{1}{0}{23}{RadarGray}{opencompass-948-secbench}%
\CensusDot{2}{0}{0}{RadarBlue}{artificial-analysis-aa-analystagent}%
\CensusDot{2}{1}{0}{RadarBlue}{artificial-analysis-aa-briefcase}%
\CensusDot{2}{2}{0}{RadarBlue}{artificial-analysis-aa-lcr}%
\CensusDot{2}{3}{0}{RadarBlue}{artificial-analysis-aa-omniscience-accuracy}%
\CensusDot{2}{4}{0}{RadarBlue}{artificial-analysis-aa-omniscience-non-hallucination}%
\CensusDot{2}{5}{0}{RadarBlue}{artificial-analysis-aime-2025}%
\CensusDot{2}{6}{0}{RadarBlue}{artificial-analysis-apex-agents-aa}%
\CensusDot{2}{7}{0}{RadarBlue}{artificial-analysis-automationbench-aa}%
\CensusDot{2}{8}{0}{RadarBlue}{artificial-analysis-critpt}%
\CensusDot{2}{9}{0}{RadarBlue}{artificial-analysis-enterpriseops-gym-aa}%
\CensusDot{2}{10}{0}{RadarBlue}{artificial-analysis-gdpval-aa-v2-normalized-score}%
\CensusDot{2}{11}{0}{RadarBlue}{artificial-analysis-gdpval-aa-v2-raw-elo}%
\CensusDot{2}{12}{0}{RadarBlue}{artificial-analysis-gpqa-diamond}%
\CensusDot{2}{13}{0}{RadarBlue}{artificial-analysis-harvey-lab-aa}%
\CensusDot{2}{14}{0}{RadarBlue}{artificial-analysis-humanitys-last-exam}%
\CensusDot{2}{15}{0}{RadarBlue}{artificial-analysis-ifbench}%
\CensusDot{2}{16}{0}{RadarBlue}{artificial-analysis-itbench-aa}%
\CensusDot{2}{17}{0}{RadarBlue}{artificial-analysis-livecodebench}%
\CensusDot{2}{18}{0}{RadarBlue}{artificial-analysis-mlcr-aa}%
\CensusDot{2}{19}{0}{RadarBlue}{artificial-analysis-mmmu-pro}%
\CensusDot{2}{0}{1}{RadarBlue}{artificial-analysis-scicode}%
\CensusDot{2}{1}{1}{RadarBlue}{artificial-analysis-tau2-bench-telecom}%
\CensusDot{2}{2}{1}{RadarBlue}{artificial-analysis-tau3-banking}%
\CensusDot{2}{3}{1}{RadarBlue}{artificial-analysis-terminal-bench-hard}%
\CensusDot{2}{4}{1}{RadarBlue}{artificial-analysis-terminal-bench-v2-1}%
\CensusDot{3}{0}{0}{RadarTeal}{aa_lcr}%
\CensusDot{3}{1}{0}{RadarTeal}{agents_last_exam}%
\CensusDot{3}{2}{0}{RadarGray}{agieval}%
\CensusDot{3}{3}{0}{RadarTeal}{aider_polyglot}%
\CensusDot{3}{4}{0}{RadarTeal}{aime}%
\CensusDot{3}{5}{0}{RadarTeal}{apex_agents}%
\CensusDot{3}{6}{0}{RadarGray}{arc_agi}%
\CensusDot{3}{7}{0}{RadarTeal}{arc_agi_2}%
\CensusDot{3}{8}{0}{RadarTeal}{arc_agi_3}%
\CensusDot{3}{9}{0}{RadarTeal}{arena_hard}%
\CensusDot{3}{10}{0}{RadarTeal}{arxivmath}%
\CensusDot{3}{11}{0}{RadarGray}{asi_bench}%
\CensusDot{3}{12}{0}{RadarTeal}{automationbench}%
\CensusDot{3}{13}{0}{RadarTeal}{babyvision}%
\CensusDot{3}{14}{0}{RadarTeal}{bankertoolbench}%
\CensusDot{3}{15}{0}{RadarTeal}{bfcl}%
\CensusDot{3}{16}{0}{RadarTeal}{bigbench_extra_hard}%
\CensusDot{3}{17}{0}{RadarTeal}{biomysterybench}%
\CensusDot{3}{18}{0}{RadarGray}{blueprint_bench_2}%
\CensusDot{3}{19}{0}{RadarTeal}{brokenarxiv}%
\CensusDot{3}{0}{1}{RadarTeal}{browsecomp}%
\CensusDot{3}{1}{1}{RadarTeal}{browsecomp_zh}%
\CensusDot{3}{2}{1}{RadarTeal}{chartography}%
\CensusDot{3}{3}{1}{RadarGray}{chartqa}%
\CensusDot{3}{4}{1}{RadarTeal}{charxiv_reasoning}%
\CensusDot{3}{5}{1}{RadarGray}{chatbot_arena}%
\CensusDot{3}{6}{1}{RadarBlue}{codeforces}%
\CensusDot{3}{7}{1}{RadarTeal}{critpt}%
\CensusDot{3}{8}{1}{RadarGray}{cursor_bench}%
\CensusDot{3}{9}{1}{RadarGray}{cvebench}%
\CensusDot{3}{10}{1}{RadarTeal}{cybergym}%
\CensusDot{3}{11}{1}{RadarTeal}{deepsearchqa}%
\CensusDot{3}{12}{1}{RadarTeal}{deepswe}%
\CensusDot{3}{13}{1}{RadarGray}{docvqa}%
\CensusDot{3}{14}{1}{RadarTeal}{draco}%
\CensusDot{3}{15}{1}{RadarTeal}{drop}%
\CensusDot{3}{16}{1}{RadarGray}{exploitbench}%
\CensusDot{3}{17}{1}{RadarGray}{frontiercode}%
\CensusDot{3}{18}{1}{RadarGray}{gdp_pdf}%
\CensusDot{3}{19}{1}{RadarBlue}{gdpval}%
\CensusDot{3}{0}{2}{RadarTeal}{gpqa}%
\CensusDot{3}{1}{2}{RadarTeal}{gpqa_diamond}%
\CensusDot{3}{2}{2}{RadarTeal}{gsm8k}%
\CensusDot{3}{3}{2}{RadarTeal}{harbor_index}%
\CensusDot{3}{4}{2}{RadarGray}{healthbench}%
\CensusDot{3}{5}{2}{RadarGray}{healthbench_professional}%
\CensusDot{3}{6}{2}{RadarTeal}{hle}%
\CensusDot{3}{7}{2}{RadarTeal}{hmmt}%
\CensusDot{3}{8}{2}{RadarTeal}{horizonmath}%
\CensusDot{3}{9}{2}{RadarTeal}{humaneval}%
\CensusDot{3}{10}{2}{RadarTeal}{ifbench}%
\CensusDot{3}{11}{2}{RadarTeal}{ifeval}%
\CensusDot{3}{12}{2}{RadarTeal}{imo_answer_bench}%
\CensusDot{3}{13}{2}{RadarTeal}{jobbench}%
\CensusDot{3}{14}{2}{RadarGray}{jointavbench}%
\CensusDot{3}{15}{2}{RadarGray}{legal_agent_benchmark}%
\CensusDot{3}{16}{2}{RadarGray}{livebench}%
\CensusDot{3}{17}{2}{RadarTeal}{livecodebench}%
\CensusDot{3}{18}{2}{RadarBlue}{livecodebench_pro}%
\CensusDot{3}{19}{2}{RadarTeal}{longbench}%
\CensusDot{3}{0}{3}{RadarTeal}{math_500}%
\CensusDot{3}{1}{3}{RadarTeal}{matharena_apex_2025}%
\CensusDot{3}{2}{3}{RadarTeal}{mathvision}%
\CensusDot{3}{3}{3}{RadarTeal}{mathvista}%
\CensusDot{3}{4}{3}{RadarTeal}{mbpp}%
\CensusDot{3}{5}{3}{RadarTeal}{mcp_atlas}%
\CensusDot{3}{6}{3}{RadarTeal}{mcp_mark}%
\CensusDot{3}{7}{3}{RadarGray}{mle_bench}%
\CensusDot{3}{8}{3}{RadarTeal}{mmlu}%
\CensusDot{3}{9}{3}{RadarTeal}{mmlu_pro}%
\CensusDot{3}{10}{3}{RadarTeal}{mmlu_redux}%
\CensusDot{3}{11}{3}{RadarTeal}{mmmlu}%
\CensusDot{3}{12}{3}{RadarTeal}{mmmu}%
\CensusDot{3}{13}{3}{RadarTeal}{mmmu_pro}%
\CensusDot{3}{14}{3}{RadarTeal}{mmvu}%
\CensusDot{3}{15}{3}{RadarTeal}{mrcr}%
\CensusDot{3}{16}{3}{RadarGray}{mtob}%
\CensusDot{3}{17}{3}{RadarTeal}{multichallenge}%
\CensusDot{3}{18}{3}{RadarTeal}{mvbench}%
\CensusDot{3}{19}{3}{RadarTeal}{nl2repo}%
\CensusDot{3}{0}{4}{RadarTeal}{office_qa_pro}%
\CensusDot{3}{1}{4}{RadarTeal}{omnidocbench}%
\CensusDot{3}{2}{4}{RadarGray}{omnivideobench}%
\CensusDot{3}{3}{4}{RadarTeal}{onemillionbench}%
\CensusDot{3}{4}{4}{RadarTeal}{osworld}%
\CensusDot{3}{5}{4}{RadarTeal}{posttrainbench}%
\CensusDot{3}{6}{4}{RadarTeal}{programbench}%
\CensusDot{3}{7}{4}{RadarGray}{rsi_bench}%
\CensusDot{3}{8}{4}{RadarTeal}{scicode}%
\CensusDot{3}{9}{4}{RadarTeal}{seccodebench}%
\CensusDot{3}{10}{4}{RadarTeal}{simpleqa}%
\CensusDot{3}{11}{4}{RadarTeal}{skillsbench}%
\CensusDot{3}{12}{4}{RadarTeal}{super_gpqa}%
\CensusDot{3}{13}{4}{RadarTeal}{superchem}%
\CensusDot{3}{14}{4}{RadarTeal}{swe_atlas}%
\CensusDot{3}{15}{4}{RadarTeal}{swe_bench_multilingual}%
\CensusDot{3}{16}{4}{RadarTeal}{swe_bench_pro}%
\CensusDot{3}{17}{4}{RadarGray}{swe_bench_science}%
\CensusDot{3}{18}{4}{RadarTeal}{swe_bench_verified}%
\CensusDot{3}{19}{4}{RadarTeal}{swe_marathon}%
\CensusDot{3}{0}{5}{RadarTeal}{tau2_bench}%
\CensusDot{3}{1}{5}{RadarGray}{tau_bench}%
\CensusDot{3}{2}{5}{RadarTeal}{terminal_bench}%
\CensusDot{3}{3}{5}{RadarTeal}{tool_decathlon}%
\CensusDot{3}{4}{5}{RadarTeal}{toolathlon_verified}%
\CensusDot{3}{5}{5}{RadarBlue}{vending_bench}%
\CensusDot{3}{6}{5}{RadarGray}{vibench}%
\CensusDot{3}{7}{5}{RadarTeal}{video_mme}%
\CensusDot{3}{8}{5}{RadarTeal}{widesearch}%
\CensusDot{3}{9}{5}{RadarTeal}{workspacebench}%
}

\newcommand{\FindingsRecords}{1283}
\newcommand{\FindingsMissingDocuments}{5}
\newcommand{\FindingsMissingRelease}{668}
\newcommand{\FindingsModelDateScores}{12594}
\newcommand{\FindingsDocumentDateScores}{322}
\newcommand{\FindingsUndatedScores}{0}
\newcommand{\FindingsTopFiveObservations}{9743}
\newcommand{\FindingsTopFiveShare}{88.0}
\newcommand{\FindingsBroadModelRecords}{22}
\newcommand{\FindingsDocumentsUnknownOrganization}{1171}
\newcommand{\FindingsRunFetched}{1003}
\newcommand{\FindingsRunDeduplicated}{954}
\newcommand{\FindingsRunPublished}{366}
\newcommand{\FindingsRunRecommended}{138}
\newcommand{\FindingsConnectorArxiv}{46}
\newcommand{\FindingsConnectorHuggingface}{101}
\newcommand{\FindingsConnectorGithub}{300}
\newcommand{\FindingsConnectorGithubOrganizations}{6}
\newcommand{\FindingsConnectorHuggingfacePapers}{26}
\newcommand{\FindingsConnectorKaggleDatasets}{30}
\newcommand{\FindingsConnectorZenodo}{65}
\newcommand{\FindingsConnectorCrossref}{180}

\newcommand{\FindingsConnectorGithubReleases}{1}

\newcommand{\FindingsConnectorOpenalex}{248}

\newcommand{\FindingsGpqaAAModels}{586}
\newcommand{\FindingsGpqaAADocuments}{1}

\newcommand{\FindingsGpqaReportsModels}{19}
\newcommand{\FindingsGpqaReportsDocuments}{27}
\newcommand{\FindingsGpqaReportsScores}{21}
\newcommand{\FindingsHleAAModels}{577}

\newcommand{\FindingsHleAAScores}{577}
\newcommand{\FindingsSciCodeAAModels}{577}

\newcommand{\FindingsCritPtAAModels}{492}

\newcommand{\FindingsHleDisplayMax}{55.47}
\newcommand{\FindingsModelDocumentRows}{%
GPQA Diamond & Artificial Analysis & 586 & 1 & 586\\
GPQA Diamond & Model reports & 19 & 27 & 21\\
Humanity’s Last Exam & Artificial Analysis & 577 & 1 & 577\\
SciCode & Artificial Analysis & 577 & 1 & 577\\
CritPt & Artificial Analysis & 492 & 1 & 492\\
}
\newcommand{\FindingsDocumentRows}{%
LLM Stats & 687 & 687 & 687 & 687\\
OpenCompass Hub & 461 & 461 & 461 & 461\\
Artificial Analysis & 25 & 25 & 23 & 23\\
Model reports & 110 & 105 & 37 & 0\\
}
\newcommand{\FindingsScoreRows}{%
LLM Stats & 687 & 679 & 8 & 5,544 & 3 & 239\\
OpenCompass Hub & 461 & 0 & 461 & 0 & Unknown & Unknown\\
Artificial Analysis & 25 & 25 & 0 & 7,050 & 245 & 586\\
Model reports & 110 & 86 & 24 & 322 & 2 & 19\\
}
\newcommand{\FindingsDateRows}{%
LLM Stats & 687 & 42 & 0 & 5,544 & 0\\
OpenCompass Hub & 461 & 457 & 0 & 0 & 0\\
Artificial Analysis & 25 & 17 & 0 & 7,050 & 0\\
Model reports & 110 & 99 & 322 & 0 & 0\\
}

\newcommand{\TaxonomyLevelOneCount}{11}
\newcommand{\TaxonomyAgenticPrimary}{128}
\newcommand{\TaxonomyAgenticCoding}{117}
\newcommand{\TaxonomyAgenticOtherClasses}{6}
\newcommand{\TaxonomyLevelTwoCount}{63}
\newcommand{\TaxonomyAgentic}{345}

\newcommand{\TaxonomyUnclassified}{4}
\newcommand{\TaxonomyLabelledFromTags}{1279}

\newcommand{\TaxonomyMinYearRecords}{30}
\newcommand{\TaxonomyExcludedFromShares}{108}
\newcommand{\TaxonomyMixAdjustment}{1.1}

\newcommand{\ReportNumber}[1]{\pgfmathprintnumber[fixed,precision=0,1000 sep={,},assume math mode=true]{#1}}
\newcommand{\datacutoff}{\ReportDataCutoff}

\graphicspath{{figures/}}

\newcommand{\RadarFigure}[4][\linewidth]{%
  \begin{figure}[!htbp]
    \centering
    \includegraphics[width=#1]{#2}
    \caption{#3}
    \label{#4}
  \end{figure}%
}
\newcommand{\RadarLinkedFigure}[5][\linewidth]{%
  \begin{figure}[!htbp]
    \centering
    \href{#2}{\includegraphics[width=#1]{#3}}
    \caption{#4}
    \label{#5}
  \end{figure}%
}
\newcommand{\RadarInlineLinkedFigure}[5][\linewidth]{%
  \begin{figure}[!htbp]
    \centering
    \href{#2}{\includegraphics[width=#1]{#3}}
    \caption{#4}
    \label{#5}
  \end{figure}%
}

\title{\vspace{-2em}\textbf{Benchmark Radar: A Living Database and Search Engine for AI Benchmarks and Evaluation}}

\renewcommand{\Affilfont}{\normalfont\fontsize{7.5}{8.5}\selectfont}

\makeatletter
\def\AB@affilsepx{\protect\quad\protect\Affilfont}
\makeatother
\makeatletter
\@ifundefined{lx@notemark}%
  {\newcommand{\CoFirstMark}{\protect\footnotemark[1]}}%
  {\newcommand{\CoFirstMark}{\thanks{These authors contributed equally to this work.}}}
\makeatother
\author[1,2]{Koutian~Wu\thanks{These authors contributed equally to this work.}\thanks{Corresponding author: \href{mailto:k@tacite.ai}{k@tacite.ai}}}
\author[4]{Wenjun~Zhang\CoFirstMark}
\author[3]{Junjie~Zhou}
\author[5]{Pengqian~Han}
\author[5,6]{Qiming~Bao}
\author[7]{Ergan~Shang}
\author[8]{Jiayu~Wang}
\author[9]{Junkai~Wang}
\author[10]{Wanghan~Xu}
\author[11]{Songyuanyi~Lu}
\author[12]{Yixuan~Li}
\author[13]{Rudy~Mizrahi~Celekli}
\author[14]{Lazizbek~Ravshanov}
\author[15,16]{Yaoliang~Bian}
\author[17]{Xudong~Han}
\author[18]{Lin~Shi}
\affil[1]{Earth-Space-AI}
\affil[2]{Tacite~AI}
\affil[3]{Hangzhou~Dianzi~University}
\affil[4]{The~University~of~Queensland}
\affil[5]{The~University~of~Auckland}
\affil[6]{Xtracta}
\affil[7]{Carnegie~Mellon~University}
\affil[8]{Xi'an~Jiaotong~University}
\affil[9]{Tsinghua~University}
\affil[10]{Shanghai~Jiao~Tong~University}
\affil[11]{The~University~of~Hong~Kong}
\affil[12]{Independent~Researcher}
\affil[13]{Gradia}
\affil[14]{University~of~Cincinnati}
\affil[15]{Westlake~University}
\affil[16]{Shenzhen~Loop~Area~Institute}
\affil[17]{University~of~Sussex}
\affil[18]{Cornell~Tech}
\date{\small Dashboard: \href{https://benchmark-radar.org/}{benchmark-radar.org}\quad
  Code: \href{https://github.com/ktwu01/benchmark-radar}{github.com/ktwu01/benchmark-radar}}
\fancypagestyle{radar}{%
  \fancyhf{}%
  \fancyhead[L]{\small\href{https://benchmark-radar.org/}{benchmark-radar.org}}%
  \fancyhead[R]{\small\href{https://github.com/ktwu01/benchmark-radar}{github.com/ktwu01/benchmark-radar}}%
  \fancyfoot[C]{\thepage}%
  }
\begin{document}
\maketitle
\vspace{-2.0em}
\begin{abstract}
Benchmark researchers and developers of large language models (LLMs) and other AI systems need to find relevant evaluations, locate their benchmark datasets and code, and understand the settings behind reported scores. We present Benchmark Radar, a living database and search engine for retrieval and discovery of AI benchmarks, covering LLM evaluation, agentic and tool-use benchmarks, coding, reasoning, safety, and domain-specific evaluations. The system combines daily discovery of benchmark papers, repositories, datasets, and releases with a searchable benchmark catalog, mentions in model cards and technical reports, and score histories. It retains source identities and citations so readers can inspect candidate benchmarks and their evaluation evidence. Daily discovery draws on \ReportIngestSourceCount{} sources: \ReportConnectorCount{} direct connectors and \ReportFirstPartyFeedCount{} first-party research and engineering feeds. The catalog contains \ReportNumber{\CensusRecords} source records drawn from \ReportCatalogSourceCount{} benchmark catalogs and \ReportNumber{\CensusNumericScores} numeric observations on \CensusScored{} records. We describe collection and retrieval, audit the full catalog, and examine benchmark saturation, adoption trends, and the limits of score comparisons. A worked example walks through a complete prior-art search, showing how to query the catalog and inspect benchmark evidence when designing a new evaluation. We release the web dashboard with a benchmark leaderboard, a Pareto frontier view of score against measured use, saturation and trend views, daily feeds, downloadable evidence, a command-line interface (CLI) for offline queries, and reproducible analysis.
\end{abstract}

\begin{figure}[!b]
\centering
\includegraphics[width=0.74\textwidth]{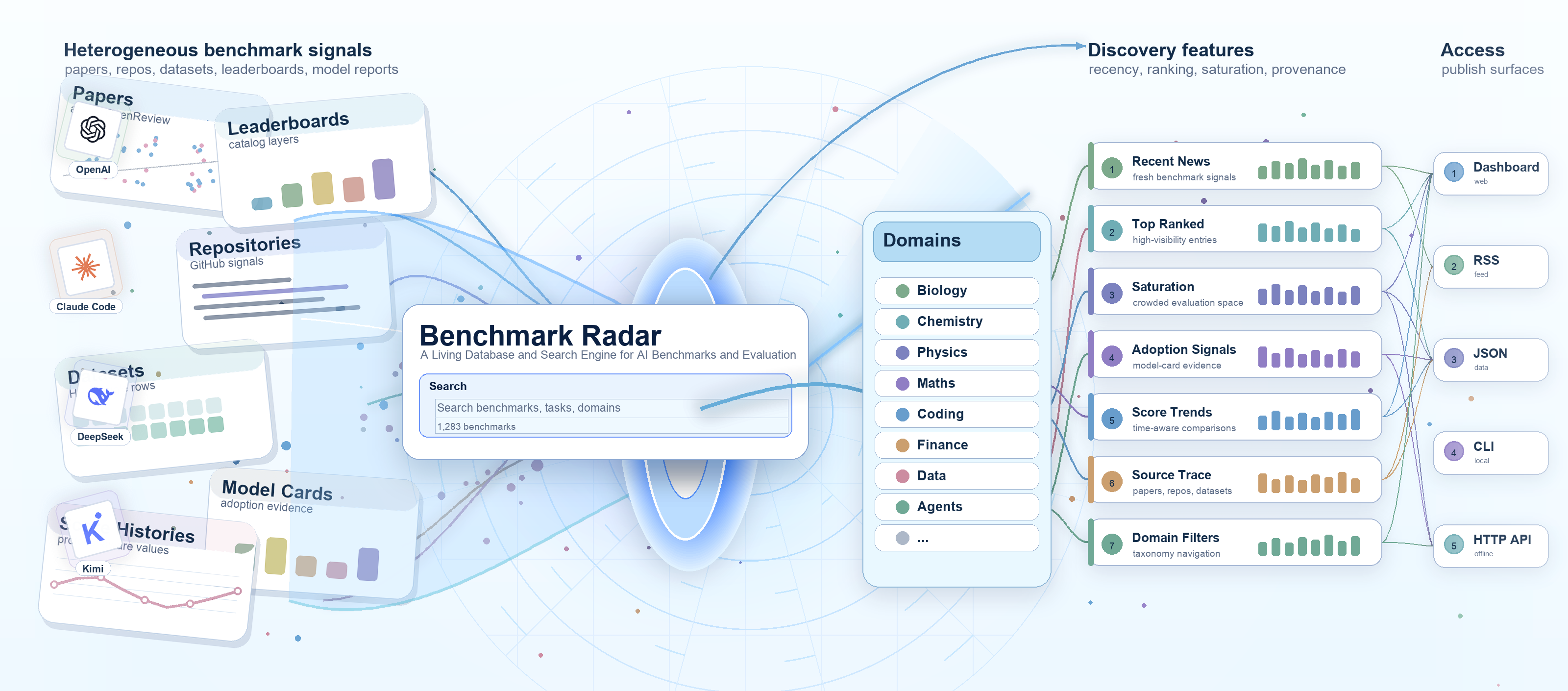}
\caption{Benchmark Radar connects discovery, retrieval, and source inspection.}
\label{fig:abstract}
\end{figure}

\clearpage
\section{Introduction}

The Transformer architecture established attention-based sequence modeling as a foundation for modern large language models (LLMs) \citep{vaswani2017attention}. GPT-6 Astra and Claude Fable 5.1 are designed for coding, research, and tasks that span multiple tools \citep{openai2026gpt6astra,anthropic2026fable51}. Comparing these models requires benchmarks that expose failures and distinguish capability gains from changes in prompts, data splits, or evaluation procedures \citep{hendrycks2021mmlu,rein2023gpqa,phan2025hle,shang2026llm}.

Evaluation also matters beyond general-purpose chat. In recommender systems, Transformer-style sequence models and LLM-based methods support user, item, and preference modeling, with efficient training needed at production scale \citep{kang2018selfattentive,sun2019bert4rec,shang2026erase}. Foundation models are also used to model biological, geoscientific, and physical processes \citep{theodoris2023transfer,cui2024scgpt,shang2025predicting,zhang2025genetic,wu2024diurnal}. In social and political analysis, language models support text annotation and simulated human samples, while network methods quantify balance and polarization dynamics \citep{leskovec2010signed,gilardi2023chatgpt,argyle2023out,li2026matraix,ng2026microverse,shang2026inference}. Benchmark scores inform claims of progress across these scientific, industrial, and decision-support settings.

Benchmarks differ in the abilities they test. Broad evaluations of knowledge and reasoning include MMLU \citep{hendrycks2021mmlu}, GPQA \citep{rein2023gpqa}, and Humanity's Last Exam \citep{phan2025hle}. Others isolate more specific capabilities: SWE-bench measures whether models can resolve real GitHub issues \citep{jimenez2023swebench}, LiveCodeBench tests code generation with date-based splits to reduce contamination from training data \citep{jain2024livecodebench}, and Terminal-Bench and Long-Horizon-Terminal-Bench test command-line agents on operational tasks \citep{merrill2026terminalbench,li2026long}. The Harbor framework ports more than 80 such benchmarks to a common execution contract so that arbitrary agents can be evaluated against them, and its Harbor-Index curates 82 difficult tasks drawn from 29 of those benchmarks \citep{shi2026harbor,harborframework}. Domain-specific evaluations include FinanceBench for finance and economics \citep{islam2023financebench}, MATH for mathematics \citep{hendrycks2021math}, SciBench for scientific problem solving \citep{wang2023scibench}, LAB-Bench for biology \citep{laurent2024labbench}, and ChemBench for chemistry \citep{mirza2024chembench}. ESM-Bench tests agents' understanding of Earth system model physics and code \citep{wu2026esm}; ResearchClawBench evaluates end-to-end autonomous scientific research \citep{xu2026researchclawbench}; and ASI-Bench examines scientific exploration and execution with progressively less methodological guidance \citep{zhou2026asi}. ScienceIDE converts scientific code repositories into executable, agent-learnable environments for training and evaluating scientific agents \citep{geng2026scienceide}. Benchmark construction is itself being automated: Benchmark Agent builds benchmarks end to end, from query analysis and subtask design to annotation and quality control \citep{xiong2026benchmark}. Their tasks and scoring rules specify which model behaviors count as successful performance.

Finding a benchmark, its dataset or repository, and reports of its use requires searching paper servers, code hosting sites, dataset hubs, vendor releases, blogs, and benchmark catalogs. Existing resources such as LLM Stats, OpenCompass, and Artificial Analysis provide leaderboards, evaluation platforms, and model-analysis views \citep{llmstats2026benchmarks,opencompass2026datasetstatistics,artificialanalysis2026methodology}. Connecting newly released benchmarks to their task materials and later use in model reports requires consulting several separate resources. Benchmark Radar addresses this gap by combining catalog retrieval with daily discovery, mentions in model reports, and scores with documented evaluation settings and sources.

Benchmark Radar gathers benchmark-related artifacts from public sources, groups observations with matching identifiers, and exposes a searchable index with source labels, dates, benchmark mentions, and scores (Figure~\ref{fig:abstract}). We present the search system, describe the collection and ranking methods needed to reproduce its outputs, and examine benchmark coverage, documentation and scored-model coverage across all catalog sources, source concentration, and gaps in evaluation settings.\footnote{Source code: \url{https://github.com/ktwu01/benchmark-radar}.}

The system preserves one record per source benchmark, including entries without scores, dates, or citations. Reviewed identity links connect related records while retaining their separate measurements. We contribute this living search system, shared web and offline access to its evidence, and a reproducible full-catalog census. Section~\ref{sec:worked-example} shows how a contributor used retrieval and source inspection to assemble prior art.

\section{Related Work and Scope}
\label{sec:related-work}

\paragraph{Benchmark catalogs and evaluations.}
LLM Stats publishes benchmark descriptions and reported model results \citep{llmstats2026benchmarks}. OpenCompass provides an evaluation platform and a benchmark registry with artifact metadata \citep{opencompass2026datasetstatistics}. Artificial Analysis publishes model evaluations and their methodology \citep{artificialanalysis2026methodology}. Benchmark Radar uses records from these sources alongside model-report evidence. Benchmark Radar adds daily discovery, artifact histories, benchmark retrieval, and source inspection through shared web and offline access, helping readers investigate evaluations across the contributing catalogs.

\paragraph{Documenting evaluation evidence.}
Model cards and datasheets motivate documenting evaluation conditions and dataset characteristics \citep{mitchell2019modelcards,gebru2021datasheets}. Benchmark Radar retains that information with its citations. Benchmarks such as GPQA \citep{rein2023gpqa}, SWE-bench \citep{jimenez2023swebench}, and SciBench \citep{wang2023scibench} define different tasks. The census measures which evidence each record provides.

\paragraph{Evaluating benchmarks themselves.}
Recent work examines benchmark saturation, item quality, and the interpretation of aggregate scores. In a systematic study of 60 language model benchmarks, \citet{akhtar2026plateau} define saturation and report that nearly half exhibit it, with prevalence increasing with benchmark age. They find that expert curation affects resilience to saturation, and that public test data has no measurable effect. Sample-level auditing of MMLU, ARC, WinoGrande, HellaSwag, and TruthfulQA identifies within-benchmark variation obscured by aggregate accuracy \citep{siedler2026benchmarks}. A reference-free judging framework assesses conversational-agent benchmarks on consistency, complexity, and policy coverage \citep{koren2026benchmarking}. For safety benchmarks developed for larger models, rankings of smaller models vary with the treatment of ambiguous responses \citep{shaik2026benchmarking}. Inverse-density weighting reduces the influence of benchmark multiplicity on aggregate scores \citep{lin2026balance}. \citet{gilda2026position} argue that evaluation scores should state their evidential scope and validity window, with conservative aggregation when component signals differ in reliability.

Benchmark Radar complements these methods by making benchmark records, available measurements, and provenance searchable. We used its command-line client to identify candidate papers for this section, supplemented the results with coauthor recommendations, and reviewed the source papers.

\section{System and Methods}

\subsection{System Overview and Daily Discovery}

Benchmark Radar adapts BuilderPulse's daily public-source collection approach \citep{builderpulse}. Figure~\ref{fig:pipeline} separates the benchmark catalog from discovery history. Model reports include model cards, technical reports, system cards, and release posts; they contribute through the same record structure as benchmark registries.

Table~\ref{tab:catalog-sources} summarizes the four catalog sources and their primary uses in the v0.11.0 release.

\begin{center}
\centering\small
\begin{tabularx}{\textwidth}{L{0.23\textwidth}r Y}
\toprule
\textbf{Catalog source} & \textbf{Rows} & \textbf{Primary use}\\
\midrule
LLM Stats & \ReportLLMStatsCount{} & Benchmark names and descriptions with source provenance.\\
OpenCompass Hub & \ReportOpenCompassCount{} & Paper, repository, release, and dataset links with source-specific identity fields.\\
Artificial Analysis & \ReportArtificialAnalysisCount{} & Current commercial evaluation catalog entries.\\
Model reports & \ReportModelReportCount{} & Benchmarks mentioned in model reports, including records without scores.\\
\midrule
Total & \ReportNumber{\ReportCatalogCount} & All four sources, retaining each record's source label.\\
\bottomrule
\end{tabularx}
\captionof{table}{Catalog sources in the v0.11.0 release. Rows count source-specific benchmark records, including records without scores.}
\label{tab:catalog-sources}
\end{center}

\subsection{Discovery Collection}

Daily discovery observations describe mentions, releases, and updates found across public sources. An observation is one collected record; an artifact is a paper, repository, dataset, release, or page linked by exact identifiers. We retain their histories alongside the catalog and count them separately from benchmark records.

Each collection run searches a 48-hour window, records counts and errors by source, removes future-dated rows, and publishes the daily snapshot once all core sources are healthy. At the cutoff, arXiv \citep{arxivapi}, Hugging Face Hub \citep{huggingfacehub}, and GitHub Search \citep{githubsearchdocs} were the core sources. Across \ReportConnectorCount{} direct connectors and \ReportFirstPartyFeedCount{} first-party feeds, additional routes cover scholarly indexes, dataset hosts, repository releases, and institutional feeds. Appendix~\ref{app:source-inventory} records their cutoff status.

\RadarFigure
  {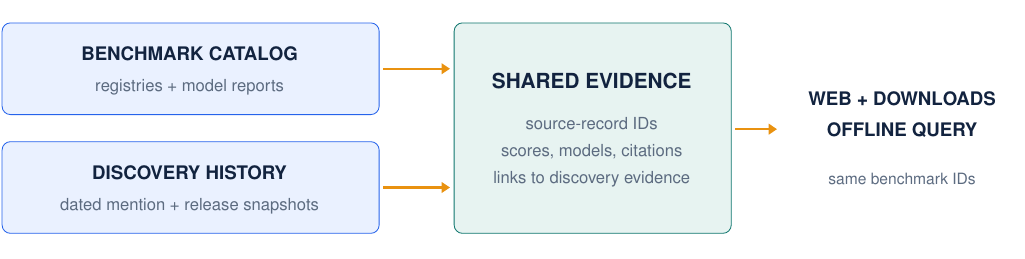}
  {Two input paths feed the published outputs: dated discovery snapshots preserve mention and release evidence, while benchmark registries and model reports populate the shared catalog. Clients use the catalog's stable source-record IDs.}
  {fig:pipeline}

Table~\ref{tab:reader-questions} describes the interfaces. Search and exports retain the full catalog; display filters apply to the current view.

\begin{center}
\centering\small
\begin{tabularx}{\textwidth}{L{0.18\textwidth}Y Y}
\toprule
\textbf{Interface} & \textbf{Reader question} & \textbf{Evidence and scope}\\
\midrule
\href{https://benchmark-radar.org/}{Today} & What appeared or changed? & Daily discovery observations, source health, and links.\\
\href{https://benchmark-radar.org/saturation}{Search} & Which benchmarks match this task? & The complete catalog, including unscored records.\\
\href{https://benchmark-radar.org/leaderboard/}{Leaderboard} & Which evaluations combine lower scores with broader measured use? & Scored source records; distinct scored models or cited documents supply the selected count.\\
\href{https://benchmark-radar.org/saturation}{Saturation} & What scores and settings were reported? & Benchmark browsing and complete score histories; search reaches every source and year.\\
\href{https://benchmark-radar.org/blog}{Blog} & Can I read and share a daily brief? & Daily briefs and an archive built from committed discovery snapshots.\\
\href{https://benchmark-radar.org/cli}{CLI} & Can I inspect the same evidence locally? & The offline query client uses the same benchmark IDs and stable JSON response format.\\
\bottomrule
\end{tabularx}
\captionof{table}{Reader questions and evidence scope.}
\label{tab:reader-questions}
\end{center}

\subsection{Preserve Records Before Comparing Measurements}

A source record is one benchmark entry from one contributing source. We normalize names, identifiers, artifact links, score observations, model identities, and cited documents into common fields. A record remains in the catalog if a score, date, or citation is absent. Reviewed identity links connect related records while preserving their separate observations and counts.

Daily discovery contributes a different kind of evidence: collected mentions, releases, and updates. Exact identifiers such as DOIs, arXiv IDs, and repository URLs link those observations to artifacts. Appendix~\ref{app:operations} records detailed collection settings and display filters; Appendix~\ref{app:source-inventory} records source health at the cutoff.

\subsection{Retrieve Candidates with Their Evidence from the Dashboard or the CLI}

The same benchmark IDs reach web search, detail pages, dataset exports, and offline clients. Lexical search uses BM25F, a field-weighted word-matching score \citep{robertson2004bm25f}, with bounded boosts for name and phrase matches. Each result exposes matched and missing query words, the fields they occur in, and the score components. A shared query service supplies the CLI and HTTP interfaces with the same response format and local data provenance.

The CLI keeps its own copy of the data. \texttt{benchmark-radar init} downloads the dataset archives and \texttt{benchmark-radar sync} updates them; a local manifest verifies each archive's SHA-256 checksum and records its schema version and provenance. Queries then run without network access, so a search can be repeated against a recorded dataset version. A search covers the normalized source records (\texttt{catalog}), the daily discovery snapshots (\texttt{radar}), or both (\texttt{all}). Filters restrict results by modality, openness, source, and the presence of a paper, repository, or dataset link. \texttt{benchmark-radar serve} answers the same queries on a local HTTP endpoint for programmatic use.

Commands that return results accept \texttt{--json}, which prints a versioned payload holding the ranking explanation above, the matching policy applied, and the version and build date of the local data. The same interface ships as an agent skill (\texttt{npx skills add ktwu01/benchmark-radar}) that states when a request calls for a benchmark query, which queries to run, and how to install the CLI and initialize its data when they are absent. A reader and an agent then work from the same records and the same score components.

An analyst or agent can try focused query variants, inspect a record's tasks and score settings, and follow its citations. The interface preserves the evidence needed to judge a candidate's suitability.

\subsection{Audit the Full Population}
\label{sec:census-method}

The census starts with every record in the rebuilt catalog index and reads its detail file. We count finite numeric score observations once by observation ID. Within each benchmark record, we count scored models by source model ID, preserving separately evaluated configurations, and cited documents by document ID. The global model count uses the registry's recorded identity links.

Eligibility depends on the measurement a calculation needs. A declared percentage unit, known score direction, and numeric values within 0--100 permit a percentage-scale summary. Percentage units come from the source's own declaration; display multipliers are recorded separately.

Date coverage counts valid recorded benchmark release dates. Score entries retain their own date basis, including model announcements and document publication; both serve as proxies for the evaluation date. Appendix~\ref{app:reproducibility} records the software revision and links to the input hashes, census script, and validation procedure.

\section{Results}
\label{sec:results}

Appendix~\ref{app:full-catalog} provides the complete linked census; Appendix~\ref{app:measurement-coverage} details document, model, score, and date coverage.

\subsection{Catalog Coverage and Task Materials}

The rebuilt catalog contains \ReportNumber{\ReportCatalogCount} source records across \ReportCatalogSourceCount{} sources. We found \ReportNumber{\CensusNumericScores} numeric score observations on \CensusScored{} records, with \CensusUnscored{} records lacking numeric scores (Table~\ref{tab:catalog-coverage}).

\RadarInlineLinkedFigure
  {https://benchmark-radar.org/leaderboard/}
  {leaderboard-frontier.png}
  {Benchmark Frontier on the Leaderboard page. The view combines reported scores, scored-model counts, and benchmark release dates or first-score date proxies. Gold rings mark the interface's Pareto candidates; hollow marks flag unverified scales or counts, and dotted outlines flag model-release date proxies. The image links to the interactive view.}
  {fig:leaderboard-frontier}

\begin{center}
\centering\small
\begin{tabular}{lrrrr}
\toprule
\textbf{Source} & \textbf{Records} & \textbf{Scored} & \textbf{Unscored} & \textbf{Numeric scores}\\
\midrule
\CensusSourceRows
\midrule
\textbf{Total} & \textbf{\ReportNumber{\CensusRecords}} & \textbf{\CensusScored} & \textbf{\CensusUnscored} & \textbf{\ReportNumber{\CensusNumericScores}}\\
\bottomrule
\end{tabular}
\captionof{table}{Score coverage across the full catalog: \CensusScored{} records have numeric observations and \CensusUnscored{} do not. Scores count numeric observations.}
\label{tab:catalog-coverage}
\end{center}

\RadarInlineLinkedFigure
  {https://benchmark-radar.org/saturation/?lscore=70&lfrontier=artificial-analysis-humanitys-last-exam}
  {hle-score-history.png}
  {Inspecting Humanity's Last Exam in Benchmark Radar. This Artificial Analysis record contains \FindingsHleAAScores{} reported scores; the highest displayed score is \FindingsHleDisplayMax{}. Marks identify model organizations, and the line connects successive best scores ordered by model release date. Release dates stand in for evaluation dates, and each score keeps the settings its source reported. The image links to the interactive view.}
  {fig:hle-scores}

Of the \CensusUnscored{} unscored records, \CensusNoScoreWithLinks{} have at least one paper, repository, or dataset link. In the full catalog, \CensusPaperLinks{} records link to papers, \CensusRepoLinks{} to repositories, and \CensusDatasetLinks{} to datasets. For example, the OpenCompass Hub record for A-Bench links a paper, code repository, and dataset despite having no archived score.

For scored records, Figure~\ref{fig:leaderboard-frontier} lets readers browse by date, reported score, and number of scored models. Figure~\ref{fig:hle-scores} shows the reported model scores within one source record.

\subsection{Benchmark Taxonomy Across the Full Catalog}
\label{sec:taxonomy}

Most source records carry a capability label. We classify \ReportNumber{\TaxonomyLabelledFromTags} of \ReportNumber{\CensusRecords} records into \TaxonomyLevelOneCount{} top-level domains and \TaxonomyLevelTwoCount{} sub-domains (Figure~\ref{fig:taxonomy-sankey}). Labels come from the publishers' own fields---OpenCompass Hub dimensions, LLM Stats categories, Artificial Analysis categories, and the model-report registry domains. The remaining \TaxonomyUnclassified{} records are LLM Stats community rows whose crawl supplied no description, category, or modality; the catalog records that as the reason.

Interaction paradigm and input modality are recorded as \emph{facets}: properties held beside a record's domain, so every record carries exactly one Level~1 class and any number of facet values. One record can count toward several facets.

Separating the axes matters: a benchmark that resolves repository issues is a coding benchmark run as an agent. Of the \TaxonomyAgentic{} agentic records, \TaxonomyAgenticPrimary{} take Agentic \& Tool Use as their Level~1 class and \TaxonomyAgenticCoding{} sit under Coding \& Software Engineering, with the rest spread across \TaxonomyAgenticOtherClasses{} further classes. A scheme with one axis has to choose, and choosing the domain hides those \TaxonomyAgenticCoding{} records from any count of agentic evaluation.

\RadarFigure[0.98\linewidth]
  {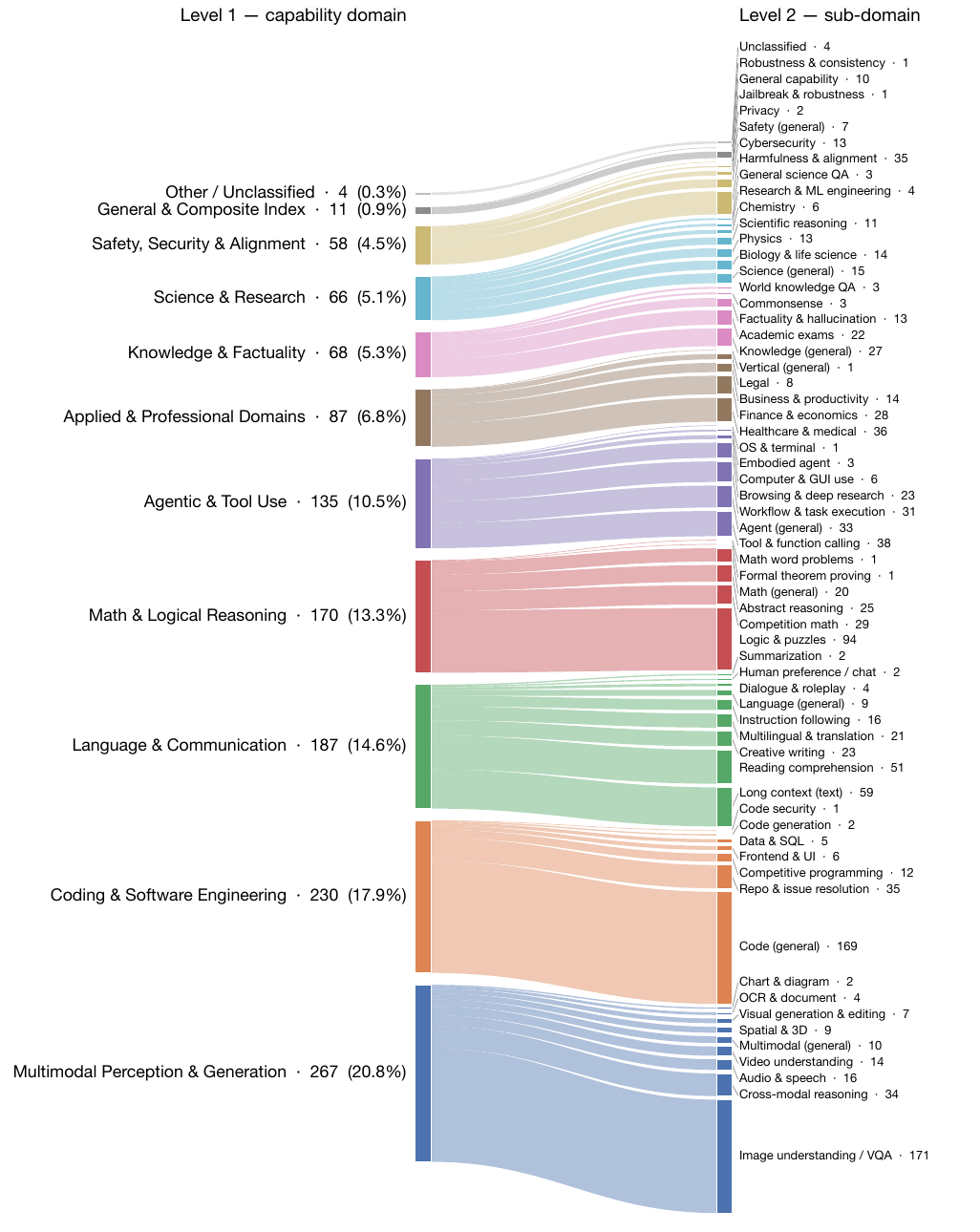}
  {Capability taxonomy over the full catalog. Ribbon height is one unit per source record, so both columns carry all \ReportNumber{\CensusRecords} records. Level~1 is one primary label per record.}
  {fig:taxonomy-sankey}

Figure~\ref{fig:taxonomy-trends} puts each of the \CensusReleaseKnown{} records with a benchmark release date on its release year. The \FindingsMissingRelease{} records without a release date stay in a separate column with the same vertical scale. Panel~B reports shares for years with at least \TaxonomyMinYearRecords{} dated records and a source mix near the pooled mix. Reweighting each of these years to a common source mix changes the agentic share by at most \TaxonomyMixAdjustment{} percentage points.

\RadarFigure
  {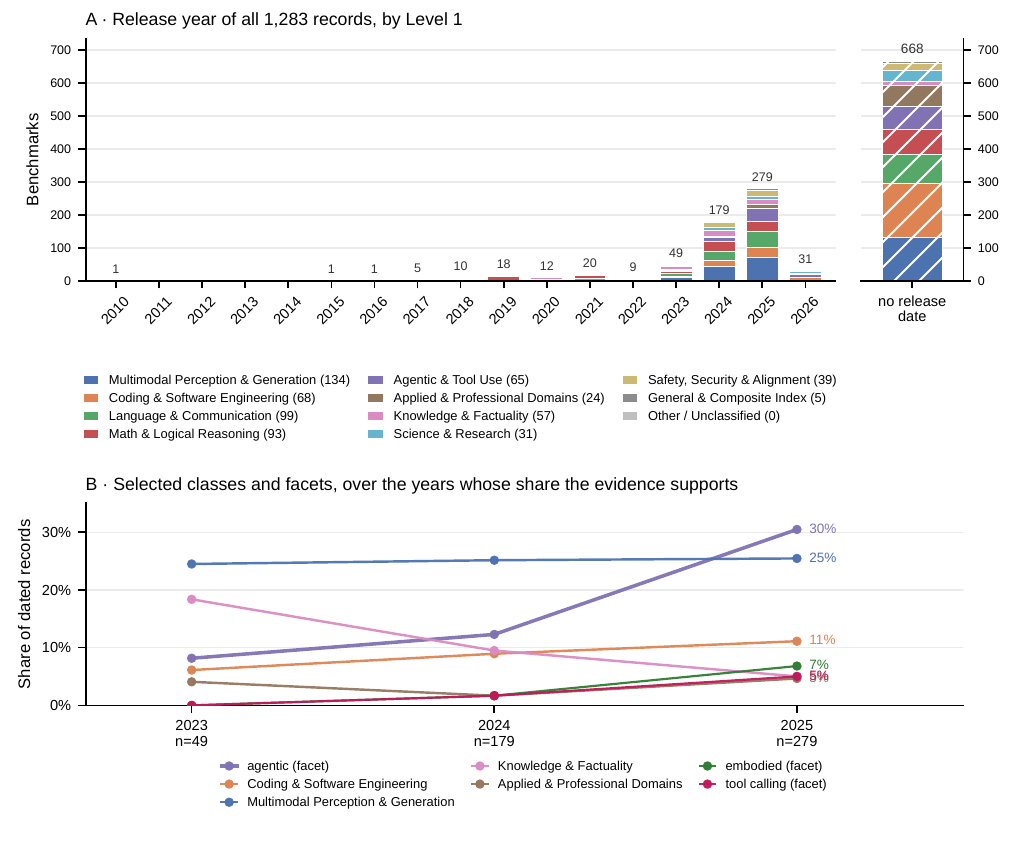}
  {Release-year composition of the classified catalog. (A)~The \CensusReleaseKnown{} dated source records by release year and Level~1 domain; the \FindingsMissingRelease{} records without a release date form the hatched column, drawn on the same scale. (B)~Shares of selected domains and facets among each year's dated records, for years with at least \TaxonomyMinYearRecords{} dated records and a source mix near the pooled mix. Panel~A also shows the \TaxonomyExcludedFromShares{} dated records from the other years. Series marked \emph{facet} cross the Level~1 classes and can overlap.}
  {fig:taxonomy-trends}

\subsection{Discovery Coverage Alongside the Catalog}

The daily discovery collection contains \ReportNumber{\ReportObservationCount} observations and \ReportNumber{\ReportArtifactCount} artifacts linked by exact identifiers across \ReportSnapshotCount{} snapshots. \CensusSimulatedSnapshots{} snapshots are simulated historical backfills; their dates describe reconstructed collection windows.

Five discovery source labels account for \ReportNumber{\FindingsTopFiveObservations} of \ReportNumber{\ReportObservationCount} observations (\FindingsTopFiveShare\%). Figure~\ref{fig:source-composition} includes the remaining labels in one aggregate bar. Source caps and collection failures shape this mix. Of the \ReportNumber{\ReportArtifactCount} artifacts, \CensusMultisourceArtifacts{} have observations from multiple sources. Cross-source matching relies on exact identifiers, so artifacts with missing or inconsistent identifiers stay unmatched.

\RadarFigure[0.92\linewidth]
  {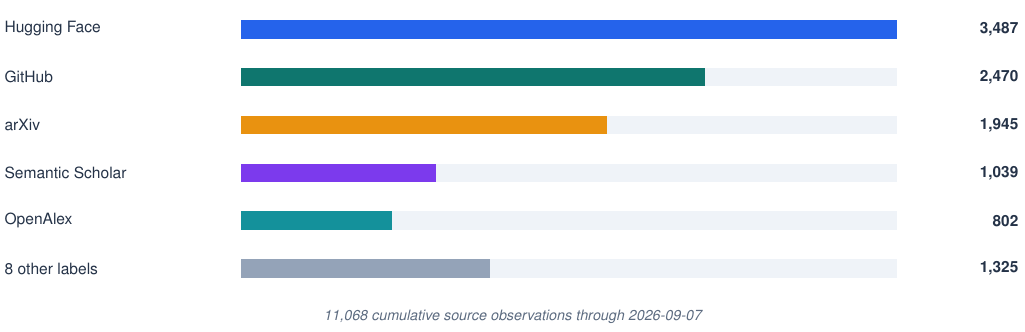}
  {Discovery observations by source through \datacutoff{}. Repeat sightings count as observations.}
  {fig:source-composition}

Figure~\ref{fig:discovery-trends} complements source coverage with the daily reading view. Its category cards separate new releases from updates, while the chart exposes changes in surfaced evidence and attention alongside collection failures.

\RadarLinkedFigure
  {https://benchmark-radar.org/trends/}
  {discovery-trends.png}
  {Discovery trends through September 7, 2026. Category cards show newly surfaced releases and track updates separately; the bars show daily evidence and attention volume. A release can carry several category tags. The visible coverage notice reports failed sources. The image links to the interactive Trends view.}
  {fig:discovery-trends}

\section{Worked Example: Checking Prior Art}
\label{sec:worked-example}

Before designing a new evaluation, a contributor searched August 2026 work on credit assignment in agentic training that provides small Qwen-series baselines for reproducible experiments. A coding agent installed the Benchmark Radar Skill and queried the catalog and the discovery layer with several topic variants. The discovery layer returned 225 matches, of which 47 matched every query word; 38 of those 47 carry an August 2026 date. After opening each candidate's paper, SRPO and HindSearch met all four conditions: an August submission, a credit-assignment method, a small Qwen backbone, and released code. Appendix~\ref{app:use-case} shows the session and the searches behind this result.

\section{Limitations and Future Work}
\label{sec:caveat-future-work}

The census describes this catalog at its recorded cutoff. Collection limits, failed requests, missing identifiers, and different snapshot dates affect its coverage. One underlying test can appear as several source records. Broader coverage requires additional source collection and review of the evidence already present.

Retrieval precision, task suitability, and time saved remain to be evaluated. The worked example uses local queries and has no controlled baseline. Lexical matching can miss paraphrases and renamed tasks. Evaluating semantic retrieval \citep{reimers2019sentencebert} will require reviewed relevance judgments across queries and candidate records, including less familiar benchmark families.

The arXiv discovery route collects new submissions only, so it misses later versions of papers first posted before the collection window. The searches used for Section~\ref{sec:related-work} missed two relevant earlier studies \citep{yang2026benchmark,akhtar2026plateau}, which a coauthor identified and read.

Measuring benchmark saturation or score stagnation requires comparable test versions and settings, with dates tied to score reporting or evaluation.

Classifying the \ReportNumber{\CensusUnclassified} discovery-derived tracks by capability, and evaluating those labels, is future work.

\section{Conclusion}

Benchmark Radar brings daily benchmark discovery, catalog search, model-report mentions, and score histories into one living search engine. Readers can find candidate evaluations, locate task materials, inspect reporting choices, and follow the settings and citations behind a score. Shared benchmark IDs connect the web dashboard, downloadable catalog, and offline query clients.

The v0.11.0 release makes \ReportNumber{\CensusRecords} source records searchable and preserves \CensusNoScoreWithLinks{} unscored records with artifact links. Its full-catalog analyses distinguish benchmark records, scored models, cited documents, and numeric observations.

\section*{Author Contributions}
\label{sec:author-contributions}

Koutian Wu and Wenjun Zhang built the collection pipeline, user interface, and implemented data aggregation across sources. Junjie Zhou designed CLI and skill of the radar. Pengqian Han contributed data analysis and copyediting. Qiming Bao contributed software, data curation, and validation. Jiayu Wang contributed a worked use case. Ergan Shang prepared figures and contributed copyediting. Songyuanyi Lu contributed discussion and improvements. Yixuan Li implemented neuroscience domain tagging for Benchmark Radar. Rudy Mizrahi Celekli, Lazizbek Ravshanov, and Yaoliang Bian contributed to software development. Junkai Wang, Wanghan Xu, and Xudong Han contributed review and copyediting. Lin Shi contributed to conceptualization and advised the work.

\section*{Acknowledgements}
\label{sec:acknowledgements}

We thank Ruohan Xu, Yifan Xu, Ran Hong, Hejia Geng, Haozhe Zhang, and Zhenxin Huang for discussion and for improving Benchmark Radar.

\begingroup
\small
\raggedright
\bibliographystyle{plainnat}
\bibliography{references}
\endgroup

\clearpage
\appendix
\begin{center}
{\LARGE\bfseries Appendix}
\end{center}
\vspace{0.75em}

\section{Full-Catalog Census}
\label{app:full-catalog}

This census retains all \ReportNumber{\CensusRecords} source records, including those without scores.

\begin{center}
\resizebox{0.98\textwidth}{!}{
\begin{tikzpicture}[x=1bp,y=1bp,every node/.style={inner sep=0pt,font=\sffamily\fontsize{8}{10}\selectfont}]
\definecolor{CensusBlue}{HTML}{2563EB}
\definecolor{CensusGray}{HTML}{94A3B8}
\definecolor{CensusInk}{HTML}{12233F}
\useasboundingbox (0,0) rectangle (492,218);
\node[anchor=west,text=CensusInk,font=\sffamily\bfseries\fontsize{12}{14}\selectfont] at (0,208)
 {\ReportNumber{\CensusRecords} records. \CensusScored{} with scores. \CensusUnscored{} without.};
\foreach \i/\label/\count in {0/{LLM Stats}/\ReportLLMStatsCount,1/{OpenCompass Hub}/\ReportOpenCompassCount,2/{Artificial Analysis}/\ReportArtificialAnalysisCount,3/{Model reports}/\ReportModelReportCount} {
 \node[anchor=west,text=CensusInk,font=\sffamily\bfseries\fontsize{8}{10}\selectfont] at (123*\i,187) {\label};
 \node[anchor=west,text=CensusInk,font=\sffamily\fontsize{7}{9}\selectfont] at (123*\i,175) {\count{} records};
}
\newcommand{\CensusDot}[5]{%
 \node[inner sep=0pt] at (2+123*#1+5.4*#2,162-3.8*#3)
  {\href{https://benchmark-radar.org/benchmarks/#5/}{\textcolor{#4}{\rule{2.6bp}{2.6bp}}}};}
\colorlet{RadarTeal}{CensusBlue}
\colorlet{RadarBlue}{CensusBlue}
\colorlet{RadarGray}{CensusGray}
\CensusMarks
\node[anchor=west,text=CensusInk,font=\sffamily\fontsize{8}{10}\selectfont] at (0,18)
 {\textcolor{CensusBlue}{\rule{5bp}{5bp}}\enspace Numeric score recorded\qquad\textcolor{CensusGray}{\rule{5bp}{5bp}}\enspace No numeric score recorded};
\node[anchor=west,text=CensusInk,font=\sffamily\fontsize{7.5}{10}\selectfont] at (0,4)
 {\CensusNoScoreWithLinks{} of the \CensusUnscored{} unscored records still link to a paper, repository, or dataset.};
\end{tikzpicture}
}
\captionof{figure}{\CensusNoScoreWithLinks{} unscored entries provide artifact links. Each mark represents one source record and links to its detail page. All \ReportNumber{\CensusRecords} records appear.}
\label{fig:corpus-evidence}
\end{center}

\section{Catalog Evidence and Measurement Coverage}
\label{app:measurement-coverage}

The following analyses use the same catalog as Section~\ref{sec:results}. They document the evidence behind its records and the requirements for comparing measurements.

\subsection{Cited Documents and Scored Models}

The common document registry contains \ReportNumber{\CensusDocuments} distinct cited documents, attached to \ReportNumber{\CensusDocumentsKnown} of the \ReportNumber{\CensusRecords} benchmark records. \FindingsMissingDocuments{} records have no cited documents. The shared model registry contains \CensusModels{} model identities. Model identifiers are available for every numeric score in all \CensusScored{} scored records, allowing distinct models to be counted within each record. For unscored records, this count is unknown.

Table~\ref{tab:model-document-coverage} compares the three counts for GPQA Diamond. The Artificial Analysis record for GPQA Diamond contains scores for \FindingsGpqaAAModels{} models and cites \FindingsGpqaAADocuments{} registry page. The Model reports record cites \FindingsGpqaReportsDocuments{} documents and contains \FindingsGpqaReportsScores{} numeric observations for \FindingsGpqaReportsModels{} models.

\begin{center}
\centering\small
\begin{tabularx}{\textwidth}{Y L{0.25\textwidth}rrr}
\toprule
\textbf{Benchmark record} & \textbf{Source} & \textbf{Models} & \textbf{Documents} & \textbf{Scores}\\
\midrule
\FindingsModelDocumentRows
\bottomrule
\end{tabularx}
\captionof{table}{One cited page can document hundreds of scored models. Models count distinct source model IDs within a record; documents count distinct citation IDs; scores count numeric observations.}
\label{tab:model-document-coverage}
\end{center}

Table~\ref{tab:model-document-coverage} also records \FindingsHleAAModels{} scored models for Humanity's Last Exam, \FindingsSciCodeAAModels{} for SciCode, and \FindingsCritPtAAModels{} for CritPt, each under one Artificial Analysis citation.

\subsection{Score Scales and Comparison Eligibility}

Only \CensusPercent{} of the \CensusScored{} scored records declare a percentage unit with a known direction and values within 0--100. The remaining \CensusOtherNumeric{} scored records use other or unverified scales. Table~\ref{tab:measurement-coverage} accounts for the entire catalog.

\begin{center}
\centering\small
\begin{tabularx}{\textwidth}{Y r L{0.46\textwidth}}
\toprule
\textbf{Measurement state} & \textbf{Records} & \textbf{Interpretation}\\
\midrule
Numeric scores, declared percentage scale & \CensusPercent & Supports scale-specific summaries.\\
Numeric scores, other or unverified scale & \CensusOtherNumeric & Scores kept on their native scale.\\
No numeric score & \CensusUnscored & Score headroom unknown.\\
\midrule
Full population & \ReportNumber{\CensusRecords} & Every source record.\\
\bottomrule
\end{tabularx}
\captionof{table}{Only \CensusPercent{} of \CensusScored{} scored records meet the percentage-scale rule in Section~\ref{sec:census-method}.}
\label{tab:measurement-coverage}
\end{center}

The catalog records benchmark release dates for \CensusReleaseKnown{} records, leaving \FindingsMissingRelease{} without a release date. Across score observations, \ReportNumber{\FindingsModelDateScores} entries carry model-announcement dates and \FindingsDocumentDateScores{} carry document-publication dates.
\newpage
\subsection{Documentation Across All Catalog Sources}

The documentation analysis covers all \ReportNumber{\FindingsRecords} benchmark records. It finds \ReportNumber{\CensusDocuments} distinct cited documents supporting \ReportNumber{\CensusDocumentsKnown} records; \FindingsMissingDocuments{} records have no cited document. Table~\ref{tab:report-mentions} covers every source. Each distinct document counts once, whether it is a registry page or a model report.

\begin{center}
\centering\small
\begin{tabularx}{\textwidth}{Y r r r r}
\toprule
\textbf{Source} & \textbf{Records} & \textbf{With docs} & \textbf{Documents} & \shortstack[r]{\textbf{Docs without}\\\textbf{named org.}}\\
\midrule
\FindingsDocumentRows
\midrule
Total & \ReportNumber{\CensusRecords} & \ReportNumber{\CensusDocumentsKnown} & \ReportNumber{\CensusDocuments} & \ReportNumber{\FindingsDocumentsUnknownOrganization}\\
\bottomrule
\end{tabularx}
\captionof{table}{Documentation across the complete v0.11.0 catalog. ``With docs'' counts benchmark records; the last two columns count distinct documents.}
\label{tab:report-mentions}
\end{center}

The release data names no organization for \ReportNumber{\FindingsDocumentsUnknownOrganization} cited documents. The complete export retains each document's identity and source URL on its benchmark record.

\subsection{Reported Scores Across the Full Catalog}
\label{sec:headroom}

The score analysis covers \ReportNumber{\CensusNumericScores} numeric observations. Table~\ref{tab:reported-headroom} summarizes scored-model coverage for each source. \FindingsBroadModelRecords{} benchmark records contain numeric scores for at least 100 distinct source model IDs.

\begin{center}
\centering\small
\begin{tabularx}{\textwidth}{Y r r r r r r}
\toprule
\textbf{Source} & \textbf{Records} & \textbf{Scored} & \textbf{Unscored} & \shortstack[r]{\textbf{Score}\\\textbf{rows}} & \shortstack[r]{\textbf{Median}\\\textbf{models}} & \shortstack[r]{\textbf{Max.}\\\textbf{models}}\\
\midrule
\FindingsScoreRows
\midrule
Total & \ReportNumber{\CensusRecords} & \CensusScored{} & \CensusUnscored{} & \ReportNumber{\CensusNumericScores} & & \\
\bottomrule
\end{tabularx}
\captionof{table}{Reported scores across every source in the v0.11.0 release. Models count distinct source model IDs within each benchmark record. Medians and maxima use records with known model counts. The record-level score export contains every catalog record, including raw maxima, all tied reporting models and citations, units, settings, and explicit missing values.}
\label{tab:reported-headroom}
\end{center}

The accompanying \texttt{evidence/catalog-findings.csv} contains one row per benchmark record, with a JSON counterpart retaining detailed evidence. Each scored record has its highest numeric value on the source's native scale and every tied score observation. Units and score direction remain attached to that record; for a lower-is-better metric, the best result is the minimum.

Percentage headroom is 100 minus a record's highest value, computed when its unit is explicitly percent, its direction is higher-is-better, and its values stay within 0--100. The export records eligibility for every benchmark. Catalog-wide headroom and recency estimates report their eligible measurement coverage alongside the statistic. Comparing results also requires matching test versions, reasoning budgets, tools, attempts, and evaluators \citep{gebru2021datasheets,mitchell2019modelcards}.

\subsection{Date Evidence Across the Full Catalog}

Table~\ref{tab:score-coverage-gaps} separates benchmark release dates from dates attached to numeric observations. Both date bases are proxies for the evaluation date.

\begin{center}
\centering\small
\begin{tabularx}{\textwidth}{Y r r r r r}
\toprule
\textbf{Source} & \textbf{Records} & \shortstack[r]{\textbf{Release}\\\textbf{known}} & \shortstack[r]{\textbf{Doc.-dated}\\\textbf{scores}} & \shortstack[r]{\textbf{Model-dated}\\\textbf{scores}} & \shortstack[r]{\textbf{Undated}\\\textbf{scores}}\\
\midrule
\FindingsDateRows
\midrule
Total & \ReportNumber{\CensusRecords} & \CensusReleaseKnown{} & \FindingsDocumentDateScores{} & \ReportNumber{\FindingsModelDateScores} & \FindingsUndatedScores{}\\
\bottomrule
\end{tabularx}
\captionof{table}{Date evidence for the complete v0.11.0 catalog. The first two numeric columns count benchmark records; the final three count numeric score observations.}
\label{tab:score-coverage-gaps}
\end{center}

The exported rows preserve each record's release date and the date basis of its scores.

\section{Collection Settings and Display Filters}
\label{app:operations}

The \datacutoff{} snapshot records \ReportNumber{\FindingsRunFetched} fetched rows and \FindingsRunDeduplicated{} candidates after duplicate removal. Of these, \FindingsRunPublished{} entered the published snapshot and \FindingsRunRecommended{} met the recommendation threshold. These counts describe the discovery run. Published evidence retains source URLs, retrieval times, parser versions, identifiers, and payload checksums. Raw responses and credentials stay private.

Benchmark Frontier, on the Leaderboard tab, requires a numeric score and excludes known pre-2024 benchmarks. It retains scored records with unknown dates or counts through labelled marks, and keeps unverified scales outside Pareto calculations. Its initial score cutoff is 70. The linked score ranking keeps its own date and numeric-score requirements.

A separate priority score for daily recommendations combines relevance (35\%), evidence (20\%), recency (20\%), and adoption (25\%) on a 0--100 scale. These weights set the reading priority of discovery observations.

\section{Source Inventory and Health}
\label{app:source-inventory}

Benchmark Radar ingests from \ReportIngestSourceCount{} sources: \ReportConnectorCount{} direct connectors and \ReportFirstPartyFeedCount{} first-party feeds. The direct connectors read arXiv \citep{arxivapi}, Hugging Face Hub and Hugging Face Papers \citep{huggingfacehub}, GitHub Search, GitHub Organizations, and GitHub Releases \citep{githubsearchdocs,githubreleasesdocs}, Kaggle \citep{kaggleapi}, Zenodo \citep{zenodoapi}, Crossref \citep{crossrefapi}, OpenAlex \citep{priem2022openalex}, OpenReview \citep{openreview}, Semantic Scholar \citep{kinney2023semanticscholar}, and Brave Search \citep{bravesearchapi}. A Hacker News collector \citep{hnalgoliaapi} supplies attention signals alongside the evidence connectors. Table~\ref{tab:healthy-connectors} lists source connectors that completed successfully at the cutoff. Row counts are measured before duplicate removal and ranking. A connector is healthy when its request succeeds and its response parses. Core sources cover papers, shared model or dataset artifacts, and code.

\begin{center}
\centering
\small
\begin{tabularx}{\textwidth}{L{0.21\textwidth}Y L{0.21\textwidth}L{0.08\textwidth}}
\toprule
\textbf{Connector} & \textbf{Role} & \textbf{Cutoff status} & \textbf{Core}\\
\midrule
arXiv & Primary papers via selected AI, language, vision, and software categories & Healthy; \FindingsConnectorArxiv{} rows & Yes\\
Hugging Face Hub & Datasets and Spaces connected to benchmark artifacts & Healthy; \FindingsConnectorHuggingface{} rows & Yes\\
GitHub Search & Code repositories and benchmark artifacts & Healthy; \FindingsConnectorGithub{} rows; at cap & Yes\\
GitHub Organizations & Reviewed organization repositories & Healthy; \FindingsConnectorGithubOrganizations{} rows & No\\
Hugging Face Papers & Community-surfaced papers & Healthy; \FindingsConnectorHuggingfacePapers{} rows & No\\
Kaggle Datasets & Public benchmark datasets & Healthy; \FindingsConnectorKaggleDatasets{} rows & No\\
Zenodo & DOI-bearing research artifacts & Healthy; \FindingsConnectorZenodo{} rows & No\\
Crossref & DOI metadata & Healthy; \FindingsConnectorCrossref{} rows & No\\
OpenReview & Conference submissions & Healthy; no eligible rows & No\\
GitHub Releases & First-party releases & Healthy; \FindingsConnectorGithubReleases{} row & No\\
OpenAlex & Scholarly discovery metadata & Healthy; \FindingsConnectorOpenalex{} rows & No\\
\bottomrule
\end{tabularx}
\captionof{table}{Healthy direct discovery connectors at the current cutoff.}
\label{tab:healthy-connectors}
\end{center}

Table~\ref{tab:first-party-feeds} lists research and engineering feeds operated by the institutions being monitored. They can surface benchmark papers, datasets, model announcements, and engineering posts. The feed collector returned no records at the cutoff. Semantic Scholar returned a malformed payload, and Brave Search lacked an API key. The snapshot records these failures. Searches restricted to an organization's website cover institutions that lack a verified feed.

\begin{center}
\centering
\small
\begin{tabularx}{\textwidth}{Y Y}
\toprule
\textbf{Feeds 1-12} & \textbf{Feeds 13-24}\\
\midrule
Meituan Engineering & Ai2\\
OpenAI News & Together AI\\
Google AI & Sakana AI\\
Google DeepMind & Qwen\\
Google Research & Ollama\\
Apple Machine Learning Research & Stability AI\\
AWS Machine Learning & Nomic AI\\
Hugging Face Blog & Replicate\\
Microsoft Research & NVIDIA Developer\\
NVIDIA AI Blog & IBM Research\\
Mistral AI & Databricks\\
Meta Research & LangChain\\
\bottomrule
\end{tabularx}
\captionof{table}{The \ReportFirstPartyFeedCount{} first-party research and engineering feeds monitored by Benchmark Radar.}
\label{tab:first-party-feeds}
\end{center}

\section{Worked Use Case: Prior-Art Check for a New Evaluation}
\label{app:use-case}

The contributor wanted to determine whether a proposed evaluation duplicated existing work. They gave the task to a coding agent together with the public setup instructions for Benchmark Radar. The agent installed the Benchmark Radar Skill and searched the catalog and discovery layers for August 2026 work on credit assignment in agentic training (Figure~\ref{fig:use-case-session}).

It compared the normalized catalog records against the unverified discovery leads and opened each August candidate's primary source to confirm the small-Qwen requirement before selecting records to compare. Figure~\ref{fig:use-case-session} lists the commands and the fields each hit returns.

\begin{center}
\resizebox{0.98\textwidth}{!}{
\begingroup
\definecolor{UCInk}{HTML}{1F2328}
\definecolor{UCMuted}{HTML}{8A8F98}
\definecolor{UCBorder}{HTML}{E3E5E8}
\definecolor{UCBg}{HTML}{FFFFFF}
\definecolor{UCPanel}{HTML}{F4F5F7}
\newcommand{\ucCmd}[1]{{\ttfamily\fontsize{6.9}{8.4}\selectfont\color{UCInk}#1}}
\newcommand{\ucBody}[1]{{\sffamily\fontsize{6.6}{7.9}\selectfont\color{UCMuted}#1}}
\newcommand{\ucLabel}[1]{{\sffamily\bfseries\fontsize{6.2}{7.6}\selectfont\color{UCMuted}#1}}
\newcommand{\ucStep}[1]{{\sffamily\bfseries\fontsize{6.6}{8}\selectfont\color{UCMuted}#1}}
\newcommand{\ucRaw}[1]{{\setlength{\fboxsep}{3pt}%
  \colorbox{UCPanel}{\begin{minipage}{\dimexpr\linewidth-2\fboxsep\relax}%
  \ttfamily\fontsize{6.3}{7.8}\selectfont\color{UCInk}#1\end{minipage}}}}
\newcommand{\ucSep}{\par\vspace{3pt}{\color{UCBorder}\rule{\linewidth}{0.5pt}}\par\vspace{3pt}}
\setlength{\fboxsep}{7pt}%
\noindent
\fcolorbox{UCBorder}{UCBg}{%
\begin{minipage}{452bp}
\setlength{\parindent}{0pt}\raggedright
\ucLabel{REQUEST}\par\vspace{2pt}
\ucBody{``Search work published in August 2026 on credit assignment in agentic training, with small Qwen-series models as reproducible baselines...''}%
\par\vspace{4pt}{\color{UCBorder}\rule{\linewidth}{0.6pt}}\par\vspace{4pt}

\begin{minipage}[t]{214bp}\raggedright
\ucStep{1}\hspace{3pt}\ucCmd{npx skills add ktwu01/benchmark-radar}\par\vspace{1.5pt}
\ucBody{skill installed; the CLI was already present}
\ucSep
\ucStep{2}\hspace{3pt}\ucCmd{benchmark-radar search "credit assignment" --scope catalog}\par\vspace{1.5pt}
\ucRaw{SCOPE: catalog   STATUS: partial\_candidates\_only\\
TOTAL\_MATCHES: 4   FULL: 0   PARTIAL: 4\\
{[2]} FraudBench   src=claire\_radar   paper=True   matched=credit missing=assignment}\par\vspace{1.5pt}
\ucBody{4 partial matches, all finance ``credit'', no benchmark record}
\ucSep
\ucLabel{QUERIES TRIED}\par\vspace{1.5pt}
\ucBody{radar: credit assignment $\cdot$ agentic credit assignment $\cdot$ step-level credit assignment $\cdot$ agentic reinforcement learning $\cdot$ Qwen}\par\vspace{1pt}
\ucBody{catalog: credit assignment $\cdot$ Qwen}
\end{minipage}%
\hspace{20bp}%
\begin{minipage}[t]{214bp}\raggedright
\ucStep{3}\hspace{3pt}\ucCmd{benchmark-radar search "credit assignment" --scope radar --limit 200}\par\vspace{1.5pt}
\ucRaw{{[27]} SRPO: Self-Reflective Policy Optimization for Long-Horizon Reasoning\\
\quad date=2026-08-25  src=arXiv  paper=True repo=False data=False\\
\quad matched=agentic+assignment+credit  missing=\\
\quad url=arxiv.org/abs/2608.23493}\par\vspace{1.5pt}
\ucBody{47 full matches of 225; 38 dated 2026-08}
\ucSep
\ucStep{4}\hspace{3pt}\ucCmd{benchmark-radar search "Qwen" --scope catalog}\par\vspace{1pt}
\ucBody{12 records, none on credit assignment. The August leads were then opened on arXiv:}\par\vspace{1.5pt}
\ucRaw{SRPO\hspace{2em}arXiv 2608.23493\hspace{2em}Qwen3-8B\\
HindSearch\hspace{1em}arXiv 2608.01597\hspace{1em}Qwen2.5-3B-Instruct}\par\vspace{1.5pt}
\ucBody{Met all four: August 2026 $\cdot$ credit assignment $\cdot$ small Qwen $\cdot$ code released}
\end{minipage}%
\end{minipage}}%
\endgroup
}
\captionof{figure}{A coding agent runs the prior-art check with Benchmark Radar. The session installs the Skill, searches the catalog and the discovery layer, and verifies each August candidate against its primary source. The \texttt{catalog} layer holds normalized benchmark records; the \texttt{radar} layer holds recent, unverified leads. Search is lexical, so each score ranks results for one query only, and every hit reports the query tokens it matched and missed together with its paper, repository, and dataset link flags.}
\label{fig:use-case-session}
\end{center}

\FloatBarrier

Table~\ref{tab:benchmark-comparison} shows a prior-art comparison table from an earlier benchmark-design effort by one contributor. Each row is one related benchmark and each column one design dimension, with each cell read from a source paper. The contributor reported that assembling it took more effort than any part of that project except producing the benchmark data. Benchmark Radar retrieves candidates for this kind of comparison and exposes the matching words and fields for inspection.

\begin{center}
\centering
\setlength{\tabcolsep}{3pt}
\renewcommand{\arraystretch}{1.2}
\centerline{\resizebox{0.98\textwidth}{!}{%
\begin{tabular}{l c c c c c c}
\toprule
\textbf{Bench Name} & \textbf{End-to-End Tasks} & \textbf{Fine-Grained Eval} & \textbf{Researcher-Quality Eval} & \textbf{Data Generation} & \textbf{Multi-Harness Eval} & \textbf{\#Tasks} \\
\midrule
MLE-Bench~\citep{chan2025mle} & \xmark & \cmark & \xmark & Transfer\&Compose & \cmark & 75 \\
MLGym-Bench~\citep{nathani2025mlgym} & \xmark & \cmark & \xmark & Automatic & \xmark & 13 \\
EXP-Bench~\citep{kon2025exp} & \cmark & \xmark & \xmark & Automatic & \cmark & 461 \\
ResearchCodeBench~\citep{hua2026researchcodebench} & \xmark & \cmark & \xmark & Transfer\&Compose & \xmark & 212 \\
MLR-Bench~\citep{chen2026mlr} & \cmark & \xmark & \xmark & Automatic & \cmark & 201 \\
PaperBench~\citep{starace2025paperbench} & \xmark & \cmark & \xmark & Transfer\&Compose & \xmark & 8316 \\
AstaBench~\citep{bragg2025astabench} & \cmark & \cmark & \xmark & Transfer\&Compose & \cmark & 2400+ \\
InnovatorBench~\citep{wu2025innovatorbench} & \cmark & \xmark & \xmark & Transfer\&Compose & \xmark & 20 \\
AIRS-Bench~\citep{lupidi2026airs} & \xmark & \cmark & \xmark & Automatic & \cmark & 20 \\
COMPOSITE-Stem~\citep{waters2026composite} & \cmark & \cmark & \xmark & Manual & \xmark & 70 \\
ScienceBoard~\citep{sun2025scienceboard} & \xmark & \cmark & \xmark & Manual & \xmark & 169 \\
\bottomrule
\end{tabular}%
}}
\captionof{table}{A prior-art comparison table assembled manually for an earlier benchmark-design effort. Each row is one related benchmark and each column one design dimension, all read from the source papers by hand.}
\label{tab:benchmark-comparison}
\end{center}

\section{Reproducibility, Access, and Citation}
\label{app:reproducibility}

All reported numbers come from \href{https://github.com/ktwu01/benchmark-radar/releases/tag/v0.13.0}{Benchmark Radar release v0.13.0}, with discovery data through \datacutoff{}. Two audit scripts recompute the census and the document, score, and date analyses from that release's catalog, registries, and dated snapshots. Capability classification and its time trends are generated separately and are not covered by these scripts.

The paper repository's \href{https://github.com/ktwu01/benchmark-radar-paper/blob/44f4146b1f67ba82b9af71d93d80b249b8887e93/README.md\#reproduce-the-papers-numbers}{reproduction guide} identifies the scripts, input hashes, and validation procedure. Table~\ref{tab:access-artifacts} lists the access and citation resources. The project also provides a Citation File Format record \citep{smith2016,katz2021,cff}.

\begin{center}
\centering
\small
\begin{tabularx}{\textwidth}{L{0.28\textwidth}Y}
\toprule
\textbf{Artifact} & \textbf{Canonical or permanent location}\\
\midrule
Source code & \url{https://github.com/ktwu01/benchmark-radar}\\
Dashboard & \url{https://benchmark-radar.org/}\\
Dataset (Hugging Face) & \url{https://huggingface.co/datasets/ktwu01/benchmark-radar}\\
RSS & \url{https://benchmark-radar.org/feed.xml}\\
Citation metadata & \url{https://github.com/ktwu01/benchmark-radar/blob/main/CITATION.cff}\\
\bottomrule
\end{tabularx}
\captionof{table}{Access and citation artifacts for Benchmark Radar.}
\label{tab:access-artifacts}
\end{center}

\section*{Licensing}
The software is released under the MIT License. This paper and the project's original editorial content are released under CC BY-NC-SA 4.0. Third-party source material remains under its original terms.

\end{document}